\documentclass[nohyperref]{article}
\usepackage{microtype}
\usepackage{graphicx}
\usepackage{subfigure}
\usepackage{booktabs} 

\usepackage{hyperref}

\usepackage[accepted]{style_icml/icml2022}

\usepackage{amsmath}
\usepackage{amssymb}
\usepackage{mathtools}
\usepackage{amsthm}

\newcommand\numberthis{\addtocounter{equation}{1}\tag{\theequation}}
\newtheorem{theorem}{Theorem}

\newtheorem{condition}{Condition}

\newtheorem{corollary}{Corollary}

\newtheorem{lemma}{Lemma}

\newtheorem{remark}{Remark}

\DeclareMathOperator{\diag}{diag}

\DeclareMathOperator{\Var}{Var}

\newcommand{\calB}{\ensuremath{\mathcal{B}}}
\newcommand{\calC}{\ensuremath{\mathcal{C}}}

\newcommand{\calG}{\ensuremath{\mathcal{G}}}

\newcommand{\calM}{\ensuremath{\mathcal{M}}}
\newcommand{\calN}{\ensuremath{\mathcal{N}}}

\newcommand{\norm}[1]{\left\|{#1}\right\|}
\newcommand{\abs}[1]{\left|{#1}\right|}

\newcommand{\set}[1]{\left\{{#1}\right\}}

\newcommand{\expec}{\ensuremath{\mathbb{E}}}

\newcommand{\prob}{\ensuremath{\mathbb{P}}}

\newcommand{\lt}{l(z^{(t)},z)}

\definecolor{asparagus}{rgb}{0.53, 0.66, 0.42}

\newcommand{\indic}{\ensuremath{\mathbf{1}}} 

\newcommand{\R}{\ensuremath{\mathbb{R}}}

\newcommand{\ircsbm}{\texttt{IR-LS}}
\newcommand{\irmap}{\texttt{IR-MAP}}
\newcommand{\lsc}{\texttt{L-SC}}
\newcommand{\ksc}{\texttt{K-SC}}
\newcommand{\orlsc}{\texttt{ORL-SC}}
\newcommand{\sdp}{\texttt{SDP-Comb}}
\newcommand{\gmm}{\texttt{EM-Emb}}
\newcommand{\asc}{\texttt{A-SC}}
\newcommand{\irmapp}{\texttt{IR-MAP(1)}}
\newcommand{\sircsbm}{\texttt{sIR-LS}}
\newcommand{\irsbm}{\texttt{IR-SBM}}
\newcommand{\kmeans}{\texttt{K-means}}
\newcommand{\sponge}{\texttt{Sponge}}
\newcommand{\irssbm}{\texttt{IR-SSBM}}
\newcommand{\ircssbm}{\texttt{IR-LSS}}
\newcommand{\spongesym}{\texttt{Sponge-sym}}

\usepackage[capitalize,noabbrev]{cleveref}

\usepackage[textsize=tiny]{todonotes}

\usepackage[font=small,skip=0pt]{caption}
\icmltitlerunning{An iterative clustering algorithm for the Contextual Stochastic Block Model with optimality guarantees}

\begin{document}

\twocolumn[
\icmltitle{An iterative clustering algorithm for the Contextual Stochastic Block Model with optimality guarantees}



\icmlsetsymbol{equal}{*}

\begin{icmlauthorlist}
\icmlauthor{Guillaume  Braun}{yyy}
\icmlauthor{Hemant Tyagi}{yyy}
\icmlauthor{Christophe Biernacki}{yyy}

\end{icmlauthorlist}

\icmlaffiliation{yyy}{Inria, Universit\'e de Lille, CNRS, Laboratoire de math\'ematiques Painlev\'e, 59650 Villeneuve d'Ascq, France}

\icmlcorrespondingauthor{Guillaume Braun}{guillaume.braun@inria.fr}

\icmlkeywords{Machine Learning, ICML}

\vskip 0.3in
]



\printAffiliationsAndNotice{}  

\begin{abstract}
Real-world networks often come with side information that can help to improve the performance of network analysis tasks  such as clustering. Despite a large number of empirical and theoretical studies conducted on network clustering methods during the past decade, the added value of side information and the methods used to incorporate it optimally in clustering algorithms are relatively less understood. We propose a new iterative algorithm to cluster networks with side information for nodes (in the form of covariates) and show that our algorithm is optimal under the Contextual Symmetric Stochastic Block Model.
Our algorithm can be applied to general Contextual Stochastic Block Models and avoids hyperparameter tuning in contrast to previously proposed methods. We confirm our theoretical results on synthetic data experiments where our algorithm significantly outperforms other methods, and show that it can also be applied to signed graphs. Finally, we demonstrate the practical interest of our method on real data.
%
\end{abstract}

\section{Introduction}
\paragraph*{}The Stochastic Block Model (SBM) is a popular generative model for random graphs -- introduced by \cite{HOLLAND1983109} -- which captures the community structures of networks often observed in the real world. Here, edges are independent Bernoulli random variables with the probability of connection between two nodes depending only on the communities to which they belong.
It is typically used as a benchmark to measure the performance of clustering algorithms. 

However, real-world networks often come with side information in the form of nodes covariates which can be used to improve clustering performance. For example, when analyzing interactions among people on a social network, we have access to additional features such as gender, age, or ethnicity that can be relevant for the clustering task. Other examples, including biological networks and predator-prey interaction networks are discussed in \cite{newman2016}. 

The Contextual Stochastic Block Model (CSBM) is a simple extension of the SBM that incorporates such  side information: each node is associated with a Gaussian vector of parameters depending only on the community to which the node belongs; see Section \ref{sec:prob_setup} for details. 

Several variants of this model and clustering algorithms have been proposed in the literature to incorporate side information. These methods include model-based approaches \citep{Yang2013CommunityDI,Weng2016CommunityDW,hric2016,emmons2019, stanley2019,contisciani2020, fontiveros2022}, spectral methods \citep{Binkiewicz2017CovariateassistedSC, Mele_2019,abbe2020ellp}, modularity based optimization methods \citep{Zhang2015CommunityDI}, belief propagation \cite{Deshpande2018ContextualSB} and semidefinite programming (SDP) based approaches \citep{CovReg}. Even if some of these algorithms come with certain theoretical guarantees, the added value of side information is in general not well understood. The recent works of \cite{abbe2020ellp}, \cite{lu2020contextual} and \cite{ma2021community} clarify the situation by establishing information theoretic thresholds for exact recovery and detection in a special case with two communities. However, the algorithm presented in \cite{abbe2020ellp} is not likely to be extended to a general CSBM with more than two (possibly unequal-sized) communities, while the latter two results focus on detection rather than consistency.

\paragraph{Our contributions.} We make the following contributions in this paper.
\begin{itemize}
    \item We propose a new iterative algorithm for clustering networks that is fast and is applicable to various settings including the general CSBM and also signed weighted graphs as shown in experiments.
    
    \item The proposed algorithm is theoretically analyzed under the CSBM, and we show that its rate of convergence is statistically optimal under the Contextual Symmetric SBM (CSSBM). As a byproduct, we derive the threshold for exact recovery with $K$ communities under the CSSBM, thus extending the recent result of \cite{abbe2020ellp} which was obtained for $K=2$. 
    
    \item We confirm the theoretical properties of our algorithm through experiments on simulated data showing that our method outperforms existing algorithms, not only under the CSBM but also under the Signed SBM the latter of which models community structure in signed networks \citep{Cucuringu2019SPONGEAG}. Finally, we provided a real data application of our algorithm. 
\end{itemize}

\paragraph{Related work.} 
As outlined earlier, covariate-assisted clustering methods have been studied from various perspectives and it is outside the scope of this work to provide an exhaustive survey. Here, we will discuss the literature that is most relevant to our work.

Our iterative method can be thought of as a Classification-EM algorithm \citep{Celeux1992ACE},  hereafter referred to as C-EM, where instead of using the likelihood we use a least squares criterion. Such ideas were first applied and analyzed under various models  including associative SBM by \cite{Lu2016StatisticalAC} and then extended to a general framework by \cite{Gao2019IterativeAF}. Recently, such ideas were also successfully applied to the  Gaussian Tensor Block Model \citep{Han2020ExactCI} and a general Gaussian Mixture Model (GMM) \citep{Chen2021OptimalCI}. However, the above results can not be  directly applied to the CSBM due to the heterogeneity of the data. 

Iterative refinement methods can also be derived naturally from the Power Method \citep{wang21pm,Ndaoud2019ImprovedCA} or alternative optimization methods \citep{NCOpt2019}. They have been  successfully deployed in other settings as well, e.g., SBM \cite{Wang2021OptimalNE}, group synchronization \cite{boumal2016}, joint alignment from pairwise differences \cite{chen2016_alignment}, graph matching \cite{gpmmatching} and low-rank matrix recovery \cite{chi2019}.

Despite the huge amount of work on covariate-assisted clustering, there are only limited consistency results.  \cite{Binkiewicz2017CovariateassistedSC} and \cite{CovReg} obtained some weak consistency guarantees that depend on both sources of information, but as noted in \cite{abbe2020ellp}, those bounds are not tight. \cite{abbe2020ellp} were the first to propose a method that achieves the threshold for exact recovery, but their algorithm only works when $K=2$ and it seems difficult to extend it to a more general setting.

\paragraph{Notation.} 
We use lowercase letters ($\epsilon, a, b, \ldots$) to denote scalars and vectors, except for universal constants that will be denoted by $c_1, c_2, \ldots$ for lower bounds, and $C_1, C_2, \ldots $ for upper bounds and some random variables. We will sometimes use the notation $a_n\lesssim b_n$ (or $a_n\gtrsim b_n$ ) for sequences $(a_n)_{n \geq 1}$ and $(b_n)_{n \geq 1}$ if there is a constant $C>0$ such that $a_n \leq C b_n$ (resp. $a_n \geq C b_n$) for all $n$. If $a_n \lesssim b_n$ and $a_n \gtrsim b_n$, then we write $a_n \asymp b_n$.  Matrices will be denoted by uppercase letters. The $i$-th row of a matrix $A$ will be denoted as $A_{i:}$ and depending on the context can be interpreted as a column vector. The column $j$ of $A$ will be denoted by $A_{:j}$, and the $(i,j)$th entry by $A_{ij}$. The transpose of $A$ is denoted by $A^\top$ and $A_{:j}^\top$ corresponds to the $j$th row of $A^\top$ by convention. $I_k$ denotes the $k\times k$ identity matrix. For matrices, we use $|| . ||$ and $||.||_F$ to respectively denote the spectral norm (or Euclidean norm in case of vectors) and Frobenius norm.  The number of non zero entries of a matrix $A$ is denoted $nnz(A)$.

\section{The statistical framework} \label{sec:prob_setup}

The CSBM consists of a graph encoded in an adjacency matrix $A\in \lbrace 0,1 \rbrace ^{n\times n}$ and nodes covariates forming a matrix $X =[X_1 \ X_2 \cdots X_n]^\top \in \mathbb{R}^{n\times d}$ where $d$ is the dimension of the covariate space. The graph and the covariates are generated as follows.

\textbf{The graph} part of the data is generated from a Stochastic Block Model (SBM) which is defined by the following parameters.  

    \noindent - The set of nodes $\calN = [n]$.
    
    \noindent - Communities $\calC_1, \ldots ,\calC_K$, of respective sizes $n_1,\ldots ,n_K$, forming a partition of $\calN$.
    
    \noindent - A membership matrix $Z \in \calM_{n,K}$ where $\calM_{n,K}$ denotes the class of membership matrices. Here, $Z_{ik}=1$ if node $i$ belongs to $\calC_k$, and is $0$ otherwise. Each membership matrix $Z$ can be associated bijectively with a function $z:[n]\to [K]$ such that $z(i)=z_i=k$ where $k$ is the unique column index satisfying $Z_{ik}=1$. To each matrix $Z \in \mathcal{M}_{n,K}$ we can associate a matrix $W$ by normalizing the columns of $Z$ in the $\ell_1$ norm: $W=ZD^{-1}$ where $D= \diag (n_1, \ldots, n_K)$.  This implies that $W^\top Z =I_K= Z^\top W$. 
    
    \noindent -  A symmetric, connectivity matrix of probabilities between communities $$\Pi=(\pi_{k k'})_{k, k' \in [K]} \in [0,1]^{K \times K}.$$ 

%
We additionally assume that the communities are approximately well balanced, i.e., \[ \frac{n}{\alpha K}\leq n_k\leq \frac{\alpha n}{K} \quad \forall k\in [K], \] for some constant $\alpha>1$. 
Denoting $P=(p_{ij})_{i,j \in [n] }:=Z\Pi Z^T$, a graph $\calG$ is distributed according to SBM$(Z, \Pi)$ if the entries of the corresponding symmetric adjacency matrix $A$ are generated by 
\[ 
A_{ij}  \overset{\text{ind.}}{\sim} \mathcal{B}(p_{ij}), \quad 1 \leq i \leq j \leq n,
\] 
where $\calB(p)$ denotes a Bernoulli distribution with parameter $p$. Hence the probability that two nodes are connected depends only on their community memberships. We will frequently use the notation $E$ for the centered noise matrix defined as $E_{ij}=A_{ij}-p_{ij}$, and denote the maximum entry of $P$ by $p_{max} = \max_{i,j} p_{ij}$. The latter can be interpreted as the sparsity level of the graph. 
We will assume throughout that $p_{max}\asymp \log n /n$. 
If the graph is denser, we are in the exact recovery regime and the problem is easy. If we are in a sparser regime, we would need to regularize the adjacency matrix to enforce concentration, but we prefer to avoid this additional technical difficulty.

For the analysis, we will also consider a special case of the SBM where the communities are equal sized, i.e., $n_k=n/K$ for all $k\in [K]$, and the connectivity matrix is given by 
\[ \Pi = 
\begin{pmatrix}
p & q & \ldots & q \\
q & p & \ldots & q \\
\vdots & \vdots & \ddots & \vdots \\
q & q & \ldots & p
\end{pmatrix} \in [0,1]^{K\times K}.
\] 
We will further assume that $p= p'\frac{\log n}{n}$ and $ q= q'\frac{\log n}{n}$ for constants $p', q'$ such that $p'>q'>0$. This model will be referred to as the Symmetric SBM and denoted by $\text{SSBM}(p,q,n,K)$.

The \textbf{nodes covariates} are generated by a Gaussian Mixture Model (GMM), independent of $A$ conditionally on the partition $Z$. 
More formally, for each $i$, \[ X_i = \mu_{z_i}+\epsilon_i, \, \text{where }\epsilon_i \overset{\text{ind.}}{\sim} \mathcal{N}(0, \sigma^2I_d) \] with $\mu_k \in \mathbb{R}^d$ for all $k\leq K$ and $\sigma >0$. We assume that $d=O(n)$.

\begin{remark}
For ease of exposition we further assume that $\sigma$ is known but our method can be extended to anisotropic GMM with unknown variance as in \cite{Chen2021OptimalCI}. We also assume $K$ to be known -- this assumption is common in the clustering  literature. Estimating $K$ is a non-trivial task which is outside the scope of this work, see \cite{Jin2021OptimalAO} for a procedure for SBM. We also leave as further work the incorporation of other forms of covariates, e.g., discrete covariates as in  \cite{Ahn2018BinaryRE}. 
\end{remark}

The \textbf{misclustering rate} associated to an estimated partition $\hat{z}$ quantifies the number of nodes assigned to a wrong cluster and is formally defined by \[ r(\hat{z},z)=\frac{1}{n}\min _{\pi \in \mathfrak{S}}\sum_{i\in [n]} \indic_{\lbrace \hat{z}(i)\neq \pi(z(i))\rbrace},\] where $\mathfrak{S}$ denotes the set of permutations on $[K]$. 
We say that we are in the exact recovery regime if $r(\hat{Z},Z)=0$ with probability $1-o(1)$ as $n$ tends to infinity. If $\prob(r(\hat{Z},Z)=o(1))=1-o(1)$ as $n$ tends to infinity then we are is the weak consistency regime. A more complete overview of the different types of consistency and the sparsity regimes where they occur can be found in \cite{AbbeSBM}. 
\section{How to integrate heterogeneous sources of information?}

The use of side information should intuitively help to recover clusters that are not well separated on each individual source of information. However, it is not well understood how to integrate two heterogeneous sources of information in the clustering process. Previous attempts \citep{Binkiewicz2017CovariateassistedSC,CovReg} proceed by directly aggregating the adjacency matrix and a Gram matrix (or Kernel matrix) formed by the covariates, but a lot of information can be lost in the aggregation process. Moreover, it is not clear what is the best linear combination of the two matrices. Here, we propose a different approach based on a two step algorithm (see Algorithm \ref{alg:ir-csbm}) that fully exploits all information. In the first step, we obtain a rough estimate of the model parameters from the previous estimate of the partition; the initialization methods that can be used are discussed in Section \ref{subsec:init}. Then, in the second step, we iteratively refine  the partition, as further explained in Section \ref{subsec:irstep}. In Section \ref{subsec:csbm_not_well_sep} we illustrate via experiments that Algorithm \ref{alg:ir-csbm} outperforms existing methods for cluster recovery in the setting where the clusters are insufficiently separated on a single source of information.

\begin{algorithm}[hbt!]
\caption{Iterative Refinement with Least Squares}\label{alg:ir-csbm}
\begin{flushleft}
        \textbf{Input:} $A\in \R^{n\times n}$, $X\in \R^{n\times d}$,$K\in \mathbb{N}^*$, $\sigma >0$, $Z^{(0)}\in \lbrace 0,1\rbrace ^{n\times K}$ a membership matrix and $T\geq 1$.
\end{flushleft}
        \begin{algorithmic}[1]
    \FOR{$0 \leq t \leq T-1$}    
        \STATE  Given $Z^{(t)}$, estimate the model parameters:  $n_k^{(t)} = |\calC_k^{(t)}|$, $W^{(t)}=Z^{(t)}(D^{(t)})^{-1}$ where $D^{(t)}= \diag (n_k^{(t)})_{k\in [K]}$, $\Pi^{(t)}=W^{(t)^\top}AW^{(t)}$ , and $\mu_k^{(t)} = W_{k:}^{(t) \top} X$, for all $k\leq K$.
        \STATE  Refine the partition by solving for each $i\in [n]$  \scriptsize \[  z_i^{(t+1)}=\arg\min_k ||(A_{i:}W^{(t)}-\Pi^{(t)}_{k:} ) \sqrt{\Sigma^{(t)}_k}||^2  + \frac{||X_i-\mu_k ^{(t)}||^2}{\sigma^2}\] \normalsize where \[\Sigma^{(t)}_k =
            \begin{cases}
            \diag (\frac{n_{k'}^{(t)}}{\Pi_{kk'}^{(t)}} )_{k'\in [K]}  &\text{(\ircsbm)} \\
            \frac{\min_{k'} n_{k'}^{(t)}}{\max_{k',k''}\Pi^{(t)}_{k'k''}}I_K  &\text{(\sircsbm)}\\
            \frac{n}{K(p^{(t)}-q^{(t)})} \log (\frac{p^{(t)}(1-q^{(t)})}{q^{(t)}(1-p^{(t)})})I_K  &\text{(\ircssbm)}
            \end{cases}
            \]
            with $p^{(t)}= K^{-1}\sum_{k\in [K]} \Pi^{(t)}_{kk}$ and $q^{(t)}=(K^2-K)^{-1}\sum_{k\neq k'\in [K]} \Pi^{(t)}_{kk'}$.
        \STATE Form the matrices $Z^{(t+1)}$ from $z^{(t+1)}$.
        \ENDFOR
        \end{algorithmic}
 \textbf{Output:} A partition of the nodes $Z^{(T)}$.
\end{algorithm}

\subsection{The refinement mechanism}
\label{subsec:irstep}
At each step $t$,  Algorithm \ref{alg:ir-csbm} estimates the model parameters given a current estimate of the partition ($W^{(t)}$), then updates the partition by reassigning each node to its closest community. Here, the proximity of a node $i$ to a community $k$ is measured by the distance between its estimated (graph) connectivity profile ($A_{i:}W^{(t)}$) and its covariates ($X_{i}$) to the current estimate of the community parameters ($\Pi^{(t)}_{k:}, \mu_k ^{(t)}$). Instead of using the Maximum A Priori (MAP) estimator as in C-EM algorithms, we use a least-square criterion. In a model-based perspective, this can be interpreted as a Gaussian approximation of the connectivity profile of each node. We will see later in the experiments that this doesn't lead to a loss of accuracy (see Section \ref{sec:xp}), and is also faster (see Table \ref{tab:time}).

Different variants of our algorithm are possible depending on the way the variance of each community is estimated and integrated in the criterion used for the partition refinement. The general method will be referred as \ircsbm, the simplified spherical version is denoted by \sircsbm\, and the version of the algorithm used for CSSBM is denoted by \ircssbm.

\textbf{Computational cost.} In each iteration, the complexity of estimating the parameters is $O(nnz(A)+nd)$ while that of estimating the partition is $O(nK(K+d))$. So the total cost of \ircsbm \, is $O(T(nnz(A)+nK(\max (K,d)))$. In our setting $A$ is sparse, hence $nnz(A)\asymp n\log n$.

\begin{remark}
Algorithm \ref{alg:ir-csbm} can also be used for clustering weighted signed graphs, as shown later in the experiments. Moreover, it is  interesting to note that when there are no covariates, the algorithm can be applied to graphs generated from a general SBM. This is in contrast to the iterative algorithm proposed by \cite{Lu2016StatisticalAC} that can only be applied to assortative SBMs (see appendix).
\end{remark}

\subsection{Initialization}
Different strategies can be adopted for initialization. If we assume that the communities are separated on each source of information and that the signal-to-noise ratio (SNR) is large enough to recover a sufficient proportion in each cluster, we can use a spectral method on one source of information (the graph for example). However, it is in general better to combine both sources of information. While one could use the methods proposed in \cite{CovReg} or \cite{Binkiewicz2017CovariateassistedSC} that also come with some theoretical guarantees, we instead use Algorithm \ref{alg:gmm} to initialize the partition. This algorithm will be referred to as \gmm. In our experiments, we used the package \texttt{clusterR} \citep{clusterr} for estimating the Gaussian mixture with an EM algorithm. This algorithm is fast, provides a sufficiently accurate estimate of the partition, and avoids hyperparameter tuning.

\label{subsec:init}
\begin{algorithm}[hbt!]
\caption{EM on graph embedding and covariates (\gmm)}\label{alg:gmm}
\begin{flushleft}
        \textbf{Input:} The number of communities $K$, the adjacency matrix $A$, covariates $X$.
\end{flushleft}
        \begin{algorithmic}[1]
        \STATE Compute $U_K \in \R ^{n\times K}$ the matrix formed by the eigenvectors associated with the top-$K$ eigenvalues (in absolute order) of $A$.
        \STATE Merge the columns of $U_K$ with the columns of $X$ to obtain a matrix $X'$.
        \STATE Cluster the rows of $X'$ by using an EM algorithm for GMM.
        \end{algorithmic}
 \textbf{Output:} A partition of the nodes $Z^{(0)}$.
\end{algorithm}

\section{Theoretical analysis}
In this section, we analyze the variants \sircsbm\, and \ircssbm \, of Algorithm \ref{alg:ir-csbm}. While it is possible to extend the analysis to \ircsbm, it would be considerably more technical and tedious due to its non- spherical structure. 
Hence, we will assume here that the covariance matrix $\Sigma_k^{(t)}$ in Algorithm \ref{alg:ir-csbm} has the form $\lambda^{(t)}I_k$ where $\lambda^{(t)}$ is an appropriate scalar depending on whether we use \sircsbm\, or \ircssbm. 

In Section \ref{sec:gen_fr} we will present the general principle for the analysis. Then we will specialize it for analyzing \ircssbm \, (under the CSSBM) in Section \ref{sec:cssbm}, and prove that the convergence rate obtained is optimal in Section \ref{sec:minimax}. Finally we show that the same framework can be used to bound the convergence rate of \sircsbm\, (under the CSBM) in Section \ref{sec:sircsbm}. The details of the proofs are outlined in the appendix.

\subsection{Analysis principle} \label{sec:gen_fr}

Our analysis is motivated by the general framework recently developed by \cite{Gao2019IterativeAF}, and also borrows some decomposition techniques used for analyzing Gaussian tensors from \cite{Han2020ExactCI}. However, these results are not directly applicable to our setting due to dependencies arising from symmetry in the SBM. Moreover, we need tighter control of the error terms then provided by these works.

We will assume w.l.o.g. that $\sigma=1$ (since $\sigma$ is assumed to be known in our framework) and that the permutation $\pi$ that minimizes the distance between $z^{(0)}$ and $z$ is the identity (if not, then replace $z$ by $\pi^{-1}(z)$). Hence there is no label switching ambiguity in the community labels of $z^{(t)}$ because they are determined from $z^{(0)}$. 

The first step is to analyze the event ``after one refinement step, the node $i$ will be incorrectly clustered given the current estimation of the partition $z^{(t)}$ at time $t$". This corresponds to the condition  
\[ a \neq \arg \min_k  ||X_i-\mu_k ^{(t)}||^2+ \hat{\lambda}^{(t)}||A_{i:}W^{(t)}-\Pi^{(t)}_{k:} ||^2 \] 
for a node $i$ such that $z_i=a$. One can see that this condition is equivalent to the existence of $b\in [K]\setminus a$ such that
{\small
\begin{equation*}
\underbrace{\langle \epsilon_i, \Tilde{\mu}_a-\Tilde{\mu}_b\rangle+ \lambda \langle E_{i:}W, \Tilde{\Pi}_{a:}-\Tilde{\Pi}_{b:}\rangle}_{C_i(a,b)} \leq \frac{-\Delta^2(a,b)}{2} +Err_{ib}^{(t)}.
\end{equation*}}
Here,
\begin{align*}
    \Delta^2(a,b)&= ||\mu_a-\mu_b ||^2+ \lambda ||\Pi_{a:}-\Pi_{b:} ||^2, \\
    \Tilde{\mu}_k &= X^\top W_{:a}\text{, }\Tilde{\Pi}_{k:}=W_{k:}^\top AW, \\
  \text{ and }  \lambda &= \frac{n_{min}}{p_{max}}\text{ or } \frac{n}{K(p-q)} \log \left(\frac{p(1-q)}{q(1-p)}\right)
\end{align*}
 depending whether we are analyzing \sircsbm\, or \ircssbm. Moreover,  $Err_{ib}^{(t)}$ is an error term that can be further decomposed as a sum $F_{ib}^{(t)}+G_{ib}^{(t)}+H_{ib}^{(t)}$ of different kinds of error terms which will be controlled in different ways. If we ignore the error term, we obtain the condition corresponding to having an incorrect result after one iteration starting from the ground truth partition. The errors occurring in this way will be quantified by the \textbf{ideal oracle error}  \[ \xi (\delta) = \sum_{i=1}^n \sum_{b \in [K]\backslash z_i} \Delta^2(z_i,b)\indic_{\lbrace C_i(a,b) \leq \frac{-(1-\delta)\Delta^2({z_i},b)}{2}\rbrace }.\]

Let us denote 
\[\Delta_{min}=\min_{a\neq b \in [K]}\Delta(a,b)\] 
%
to quantify the separation of the parameters associated with the different communities. If $\Delta_{min}=0$, it would imply that at least two communities are indistinguishable and the model would not be identifiable.
%
For $t \geq 1$ and $\delta \in [0,1)$, let 
\[ \delta^{(t)}= \max \left(\frac{7}{8}\frac{\tau^{(t-1)}}{\tau^{(0)}}, \delta \right), \ \tau^{(t)} = \tau^{(0)} \delta^{(t)} \] 
be sequences where $\tau^{(0)}=\epsilon n\Delta_{min}^2/K$ for a small enough constant $\epsilon>0$.  

In general the rate of decay of $\xi(\delta)$ leads to the convergence rate of iterative refinement algorithms, hence it is important to control this quantity.

\begin{condition}[ideal error]\label{cond:ideal} Assume that \[ \xi (\delta^{(t)}) \leq \frac{3}{4}\tau^{(t-1)}, \text{ for all }t\geq 1 \] holds with probability at least $1-\eta_1$.
\end{condition}

We now have to analyze the error terms and prove that their contribution is negligible compared to the ideal oracle error rate. Let  \[ l(z,z') = \sum_{i\in [n]} \Delta^2(z_i,z'_i)\indic_{\lbrace z_i\neq z_i'\rbrace } \] be a measure of distance between two partitions $z, z'\in [K]^n$. We will control the error terms by showing that the following conditions are satisfied.
\begin{condition}[F-error type]\label{cond:f}
Assume that \[ \max_{\lbrace z^{(t)}: l(z,z^{(t)})\leq \tau^{(0)} \rbrace}\sum_{i=1}^n\max_{b\in [K]\backslash z_i} \frac{(F_{ib}^{(t)})^2}{\Delta^2(z_i,b)l(z,z^{(t)})} \leq \frac{\delta^2}{256}\] for all $t \geq 0$ holds with probability at least $1-\eta_2$.
\end{condition}


\begin{condition}[GH-error type]\label{cond:h}
Assume that \[ \max_{i\in [n]}\max_{b \in [K]\setminus z_i} \frac{|H_{ib}^{(t)}|+|G_{ib}^{(t)}|}{\Delta (z_i,b)^2 } \leq \frac{\delta^{(t+1)}}{4} \] holds uniformly on the event $\lbrace z^{(t)}: l(z,z^{(t)})\leq \tau^{(t)} \rbrace$ for all $t\geq 0$ with probability at least $1-\eta_3$ .
\end{condition}

We can now show under these conditions that there is a contraction of the error if the initial estimate of the partition is close enough to the ground truth partition.

\begin{theorem}\label{thm:gao_ext}  Assume that $ l(z^{(0)},z) \leq \tau^{(0)}$ and $\delta<1$.
Additionally assume that Conditions \ref{cond:ideal}, \ref{cond:f}, and \ref{cond:h} hold. Then with probability at least $1-\sum_{i=1}^3\eta_i$ 
\begin{equation}\label{eq:contraction}
    \lt \leq \xi(\delta^{(t)}) + \frac{1}{8}l(z^{(t-1)},z), \, \forall t \geq 1.
\end{equation} 
\end{theorem}

\begin{remark}
This is an adaptation of Theorem 3.1 in \cite{Gao2019IterativeAF} where we allow at each step to choose a different $\delta^{(t)}$. It allows us to obtain a weaker condition for initialization than the one used in Theorem 4.1 in  \cite{Gao2019IterativeAF}. Indeed, they require $l(z^{(0)},z)=o(\frac{n\Delta^2_{min}}{K})$ in order to have $\delta=o(1)$, but we only need $l(z^{(0)},z) = O(\frac{n\Delta^2_{min}}{K})$.  
\end{remark}
%

\begin{proof}[Proof of Theorem \ref{thm:gao_ext}]
By definition 
$\delta^{(t)}<1$ for all $t \geq 1$.  Let $i \in [n]$ such that $z_i=a$ and assume that $l(z^{(t-1)},z) \leq \tau^{(t-1)}$ for some given $t \geq 1$. Denoting $I^{(t)}_i(a,b) := \indic_{\lbrace C_i(a,b) \leq \frac{-(1-\delta^{(t)})\Delta^2(a,b)}{2} \rbrace}$, observe that 
%
%
\begin{align*}
    &\indic_{\lbrace z_i^{(t)}=b\rbrace} \\
    & \overset{(1)}{\leq} \indic_{\lbrace C_i(a,b) \leq \frac{-\Delta^2(a,b)}{2} +F_{ib}^{(t-1)}+G_{ib}^{(t-1)}+H_{ib}^{(t-1)} \rbrace }
    \\
    &\overset{(2)}{\leq} I^{(t)}_i(a,b) 
    + \indic_{\lbrace \frac{\delta^{(t)}}{2}\Delta^2(a,b)\leq F_{ib}^{(t-1)}+G_{ib}^{(t-1)}+H_{ib}^{(t-1)}\rbrace}
    \\
    &\overset{(3)}{\leq} I^{(t)}_i(a,b)  
     + \indic_{\lbrace \frac{\delta^{(t)}}{4}\Delta^2(a,b)\leq F_{ib}^{(t-1)}\rbrace} 
     \\
    &\leq I^{(t)}_i(a,b) + \indic_{\lbrace \frac{\delta}{4}\Delta^2(a,b)\leq F_{ib}^{(t-1)}\rbrace}
    \\
    &\overset{(4)}{\leq} I^{(t)}_i(a,b) +\frac{32(F_{ib}^{(t-1)})^2}{\delta^2\Delta^4(a,b)}.
\end{align*}

The inequality $(1)$ follows from the definition of $z_i^{(t)}$ and the error decomposition. Inequality $(2)$ comes from a union bound while $(3)$ uses Condition \ref{cond:h}. Finally, $(4)$ follows from Markov inequality. Hence,
\begin{align*}
    &\lt = \sum_{i\in [n]} \sum_{b\in [K]\backslash \lbrace z_i \rbrace}\Delta^2(z_i,b)\indic_{\lbrace z_i^{(t)}=b\rbrace}\\
    & \leq \sum_{i\in [n]} \sum_{b\in [K]\backslash \lbrace z_i \rbrace}\Delta^2(z_i,b) \indic_{\lbrace C_i(a,b) \leq \frac{-(1-\delta^{(t)})\Delta^2(z_i,b)}{2} \rbrace}\\
    & +\sum_{i\in [n]} \sum_{b\in [K]\backslash \lbrace z_i \rbrace}\Delta^2(z_i,b)\indic_{\lbrace z_i^{(t)}=b\rbrace}\frac{32(F_{ib}^{(t-1)})^2}{\delta^2\Delta^4(z_i,b)} \\
    & \leq \xi(\delta^{(t)}) + \sum_{i\in [n]}  \max_{b\in [K]\backslash \lbrace z_i \rbrace}\frac{32(F_{ib}^{(t-1)})^2}{\delta^2\Delta^2(z_i,b)} \\
    &\leq \xi(\delta^{(t)}) + \frac{1}{8}l(z^{(t-1)},z).
\end{align*}
Using Condition \ref{cond:ideal}, we hence obtain
\[ \lt \leq \xi(\delta^{(t)}) +\frac{1}{8}\tau^{(t-1)} \leq \frac{7}{8}\tau^{(t-1)}.\] 
%
Thus $\tau^{(t)}$ is an upper bound for $\lt$ and the theorem is proved by induction.
\end{proof}
By iteratively unwrapping \eqref{eq:contraction} we obtain for $t$ large enough the following bound on $\lt$.
\begin{corollary}\label{cor:contraction} Under the assumptions of Theorem $1$, we have for all $t\gtrsim \log(1/\delta)$ that  \[ \lt \lesssim \xi(\delta)+ \tau^{(0)}(1/8)^{t-\Theta (\log(1/\delta))}. \] 
\end{corollary}
This bound shows that if $t$ is suitably large, then the estimation error is of the order of the oracle error.

\subsection{Convergence guarantees for \ircssbm\, under the  CSSBM}\label{sec:cssbm}
Let us define the SNR 
\begin{align*}
    \Tilde{\Delta}^2 &= \frac{1}{8}\min_{k\neq  k'} ||\mu_k-\mu_{k'} ||^2+\frac{\log n }{K}(\sqrt{p'}-\sqrt{q'})^2. 
\end{align*} 
It is easy to see that $\tilde{\Delta}\asymp \Delta_{min}$. 
The following lemma shows that $\xi(\delta)$ decreases exponentially fast in $\tilde{\Delta}$ provided $\Delta_{min}$ is suitably large.
\begin{lemma}\label{lem:ideal_error} Assume that $K^{1.5}/\Delta_{min}\to 0$ and $\delta=\delta(n) \rightarrow 0$ at a suitably slow rate. Then with probability at least $1-\exp(-\tilde{\Delta})$, we have
\[  \xi(\delta) \leq n\exp(-(1+o(1))\Tilde{\Delta}^2) .\] 
\end{lemma}
The following theorem shows that if $z^{(0)}$ is close enough to $z$, then the misclustering rate decreases exponentially fast with the SNR $\tilde{\Delta}$ after $O(\log n)$ iterations. 
\begin{theorem}\label{thm:IRcv} Assume that $K^{1.5}/\Delta_{min}\to 0$ and $\tilde{\Delta}^2\asymp \log n/K$. Under the CSSBM$(p,q,n,K)$ assumption, if $z^{(0)}$ is such that \[ l(z,z^{(0)}) \leq \frac{\epsilon n\Delta^2_{min}}{K}\] for a constant  $\epsilon$ small enough, then with probability at least $1-n^{-\Omega(1)}$ we have for all $t\gtrsim \log n$ \[ r(z^{(t)},z) \leq \exp (-(1+o(1))\tilde{\Delta}^2).\] 
\end{theorem}
\begin{proof}[ Sketch of proof.] We first show that Conditions \ref{cond:f} and \ref{cond:h} are satisfied. Then we show that Condition \ref{cond:ideal} is satisfied for the sequences $\delta^{(t)}$ and $\tau^{(t)}$, hence Theorem \ref{thm:gao_ext} can be applied to obtain a contraction of the error at each step.
\end{proof}


\begin{remark}
By assumption $\tilde{\Delta}^2 \gtrsim \log n/K$, and so
the condition $\tilde{\Delta}^2\asymp \log n/K$ is not very restrictive. Indeed, if the information provided by the GMM part was not of the same order as the  graph part, it would not be useful to aggregate information. If $\tilde{\Delta}^2\gg \log n$ then we would be in the exact recovery setting and the problem becomes easy.
\end{remark}

\begin{remark}
The initial condition implies that $h(z^{(t)},z)/n\leq \epsilon/K$ where $h$ denotes Hamming distance, see appendix for details. This is a detection condition. It is outside the scope of this work to analyze an algorithm that achieves this condition under CSBM. The numerical experiment done in the appendix suggests that the algorithm can still work with a random initialization, at least in particular cases.
\end{remark}

\subsection{Minimax lower-bound for CSSBM}\label{sec:minimax}
We are going to establish that the convergence rate established in Theorem \ref{thm:IRcv} is optimal. Let \[ \Theta = \lbrace (\mu_k)_{k \in [K]} \in \R^K, p,q \in [0,1] \text{ such that } p>q \rbrace\] be the admissible parameter space.
\begin{theorem}\label{thm:minimax} Under the assumption $\Tilde{\Delta}/\log K\to \infty$, we have \[ \inf_{\hat{z}} \sup_{\theta \in \Theta} \expec (r(\hat{z},z)) \geq \exp(-(1+o(1))\Tilde{\Delta}^2). \] If $\Tilde{\Delta} +\log K=O(1)$, then $\inf_{\hat{z}} \sup_{\theta \in \Theta} \expec (\frac{r(\hat{z},z)}{n}) \geq c$ for some positive constant $c$.
\end{theorem}
%
\begin{remark}
This lower-bound shows that if $\tilde{\Delta}^2< \log n$ then every estimator fails to achieve exact recovery with a probability bounded below from zero because $\sup_{\theta \in \Theta} \expec (r(\hat{z},z)) > n^\epsilon $ for some $\epsilon >0$. On the other hand, Theorem \ref{thm:IRcv} shows that when $\tilde{\Delta}^2> \log n$ then \ircssbm\, achieves exact recovery. Hence the threshold for exact recovery is $\tilde{\Delta}^2/\log n$. When $K=2$ and $\mu_1=-\mu_2=\mu$ this matches the result obtained by  \cite{abbe2020ellp}.
\end{remark}
\begin{proof}[Sketch of proof.] We can use the same argument as in  Theorem 3.3 \cite{Lu2016StatisticalAC} to reduce the problem to a hypothesis testing problem. The solution of the latter is given by the maximum likelihood test according to the  Neyman-Pearson lemma. Then the probability of error can be controlled by using concentration inequalities.
\end{proof}

\subsection{Convergence guarantees for \sircsbm}\label{sec:sircsbm}
The proof techniques used in the previous section can be extended in a straightforward way to obtain consistency results for \sircsbm \, under the CSBM. The main difference is that the specialized concentration inequality used to prove Lemma \ref{lem:ideal_error} can no longer be applied to this setting. 

\begin{theorem}\label{thm:sirls} Assume that $K^{1.5}/\Delta_{min}\to 0$, $\Delta_{min}^2\asymp \log n/K$ and $\max_{a,b\in [K]}\Delta^2(a,b) \lesssim \Delta_{min}^2 $. Under the CSBM with approximately balanced communities, if $  l(z,z^{(0)}) \leq \frac{\epsilon n\Delta^2_{min}}{K}$ for some small enough constant  $\epsilon>0$, then with probability at least $1-n^{-\Omega(1)}$ we have  \[ r(z^{(t)},z) \leq \exp \left(-\frac{1}{8}\Delta_{min}^2\right) \text{ for all $t\gtrsim \log n$}.\]  
\end{theorem}

\paragraph{Existing theoretical results.}
\cite{abbe2020ellp} obtain the same bound as us when $K=2$. Their method -- which is an aggregated spectral method --  requires computing the largest eigenvector of $XX^\top$ (with diagonal set to zero) which has a complexity $O(n^2)$. In contrast, our method has complexity $O(n\log n+nd)$ and is thus faster when $d=o(n)$. \cite{Binkiewicz2017CovariateassistedSC} consider a spectral method applied on a regularized Laplacian. When $d\asymp \text{polylog} (n)$ they show that the misclustering rate is $O(1/\text{polylog} (n))$. \cite{CovReg} used a similar regularization idea but with a SDP and obtain an error bound (in Frobenius norm) for estimating a clustering matrix. Their bound depends on the two sources of information, but as also noted by \cite{abbe2020ellp}, it is unclear how the bound improves with side information. Moreover the bounds in \cite{Binkiewicz2017CovariateassistedSC,CovReg} are not optimal.


\section{Numerical experiments}\label{sec:xp}
%
We now empirically evaluate our method on both synthetic and real data\footnote{The source code is available on \url{https://github.com/glmbraun/CSBM}}. Section \ref{subsec:csbm_not_well_sep} contains simulations for the CSBM and Section \ref{sec:xpsigned} contains results for clustering signed networks under a Signed SBM. In Section \ref{subsec:aus_dataset}, we test our method on a dataset consisting of a (weighted) signed graph along with covariate information for the nodes.
%
%
\subsection{CSBM with not well separated communities} \label{subsec:csbm_not_well_sep}
In this experiment the graph is generated from a SBM with parameters $n=1000$, $K=3$, $Z_i \overset{i.i.d}{\sim }\text{Multinomial}(1; 1/3,1/3,1/3)$, and \[ \Pi = 0.02*\begin{pmatrix}
                        1.6 & 1.2 & 0.05 \\
                        1.2 & 1.6 & 0.05 \\
                        0.05 & 0.05 & 1.2 
                      \end{pmatrix}.
\] The covariates are generated from a GMM with variance $\sigma^2 = 0.2$ and class centers $\mu_1 =(0,0,1),\mu_2= (-1,1,0),$ $\mu_3 = (0,0,1)$. Note that  $\calC_1,\calC_3$ cannot be separated by the covariate information, while  $\calC_1,\calC_2$ are not well separated in the graph information (as seen from $\Pi$). Hence, one would expect in this example that using only a single source of information should not yield good clustering results.  
%
To demonstrate this, we use the Normalized Mutual Information (NMI) criterion to measure the quality of the resulting clusters. It is an information theoretic measure of similarity taking values in $[0,1]$, with $1$ denoting a perfect match, and $0$ denoting the absence of correlation between partitions. 

\paragraph{Performance comparison.} We will use \ksc \ and \lsc \ to denote respectively the results obtained by applying spectral clustering on the Gaussian kernel matrix $K$ formed from the covariates, and spectral clustering (SC) applied on the Laplacian of the graph. Additionally, \sdp \ refers to the method  proposed by \cite{CovReg}; \irmap \ is similar to \ircsbm \ but with the least-square criterion replaced by the MAP to update the partition; \orlsc \ (Oracle Regularized Laplacian SC) corresponds to SC applied on $A + \lambda K$ where $\lambda$ is chosen to maximize the NMI between the (oracle known) true partition and the one obtained by using SC on  $A + \lambda K$. For the implementation of \sdp, we used the Matlab code provided by \cite{CovReg} with the $ \lambda$ given by \orlsc. We limited our comparison to the aforementioned methods for concreteness; a comparison with all existing methods from the literature would need a separate study and is outside the scope of the paper. 
%

\begin{figure}
    \centering
    \includegraphics[scale=0.45]{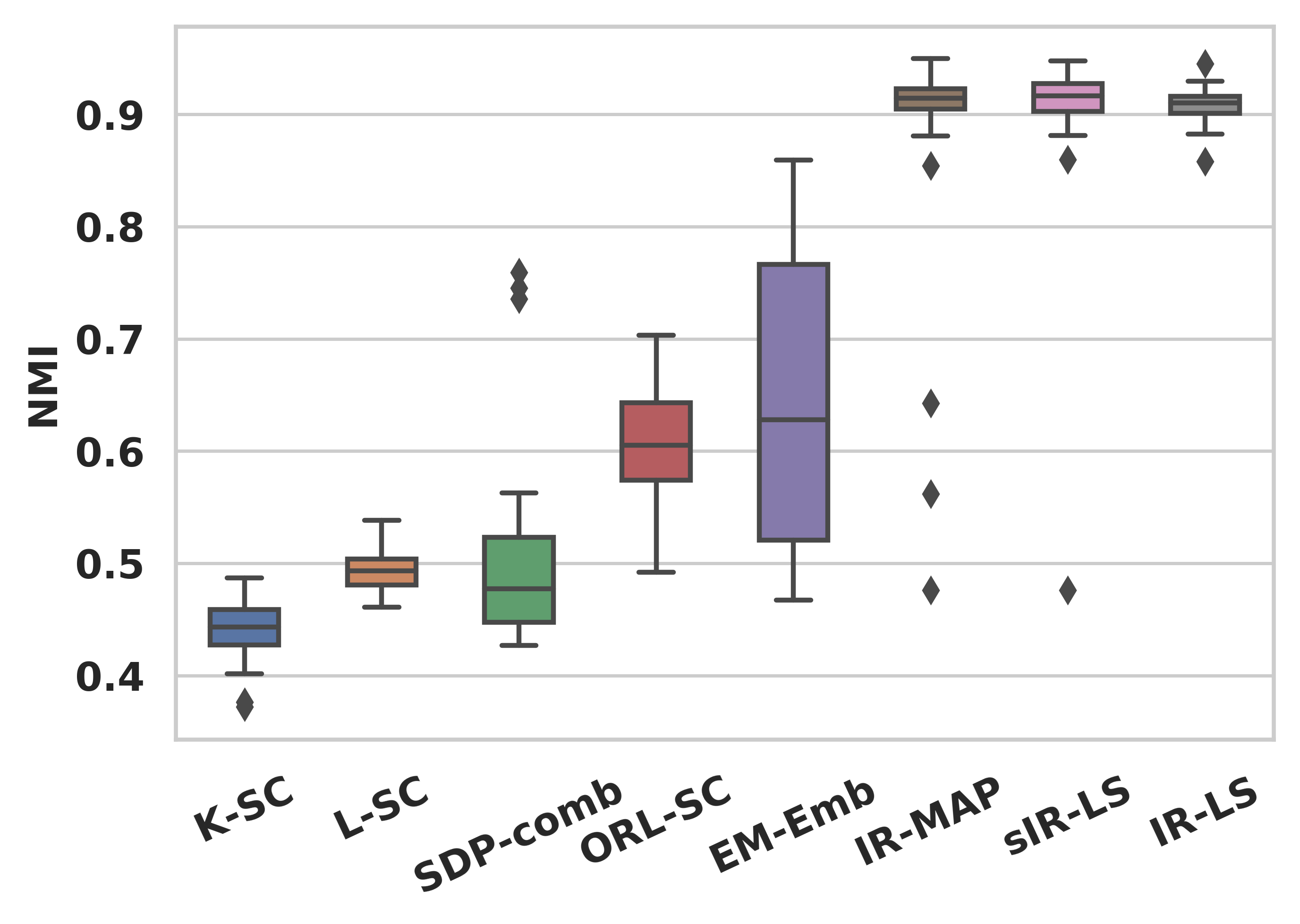}
    \caption{{\small Average performance over $40$ runs  of different algorithms under CSBM.}}
    \label{fig:boxplot}
\end{figure}

Figure \ref{fig:boxplot} shows that the three iterative methods considered (\irmap, \ircsbm, \sircsbm), initialized with \gmm, provide significantly better clustering performance compared to the other methods. 
The variance of \sircsbm\, is a bit larger than \ircsbm.
On the other hand, other methods based on  aggregating the two sources of information (\sdp\, and \orlsc\,) lead to a limited improvement in clustering performance. Additional experiment results in the appendix suggest that the iterative methods considered also work with random initialization.

\paragraph{Computational cost.} We took the average of CPU time (in seconds) over $20$ repetitions.  
There is an important gain in speed obtained by replacing the MAP objective by a least square criterion.  Moreover, the initialization obtained with \gmm\, is very fast. The results are gathered in Table \ref{tab:time}.
\begin{table}[H]
    \centering
    \resizebox{\columnwidth}{!}{
    \begin{tabular}{|c|c|c|c|c|c|}
    \hline
         & \lsc & \orlsc & \gmm & \ircsbm & \irmap \\ 
         \hline
Time& 1.4 & 7.9 & 0.5 & 1.2 & 37.1 \\
\hline
Ratio & 2.7 & 15 & 1 & 2.3 & 70 \\
\hline
    \end{tabular}}
    \caption{Comparison between computation times (averaged over $20$ runs)}
    \label{tab:time}
\end{table}

\subsection{Signed SBM}\label{sec:xpsigned}
\begin{figure}
    \centering
    \includegraphics[scale=0.38]{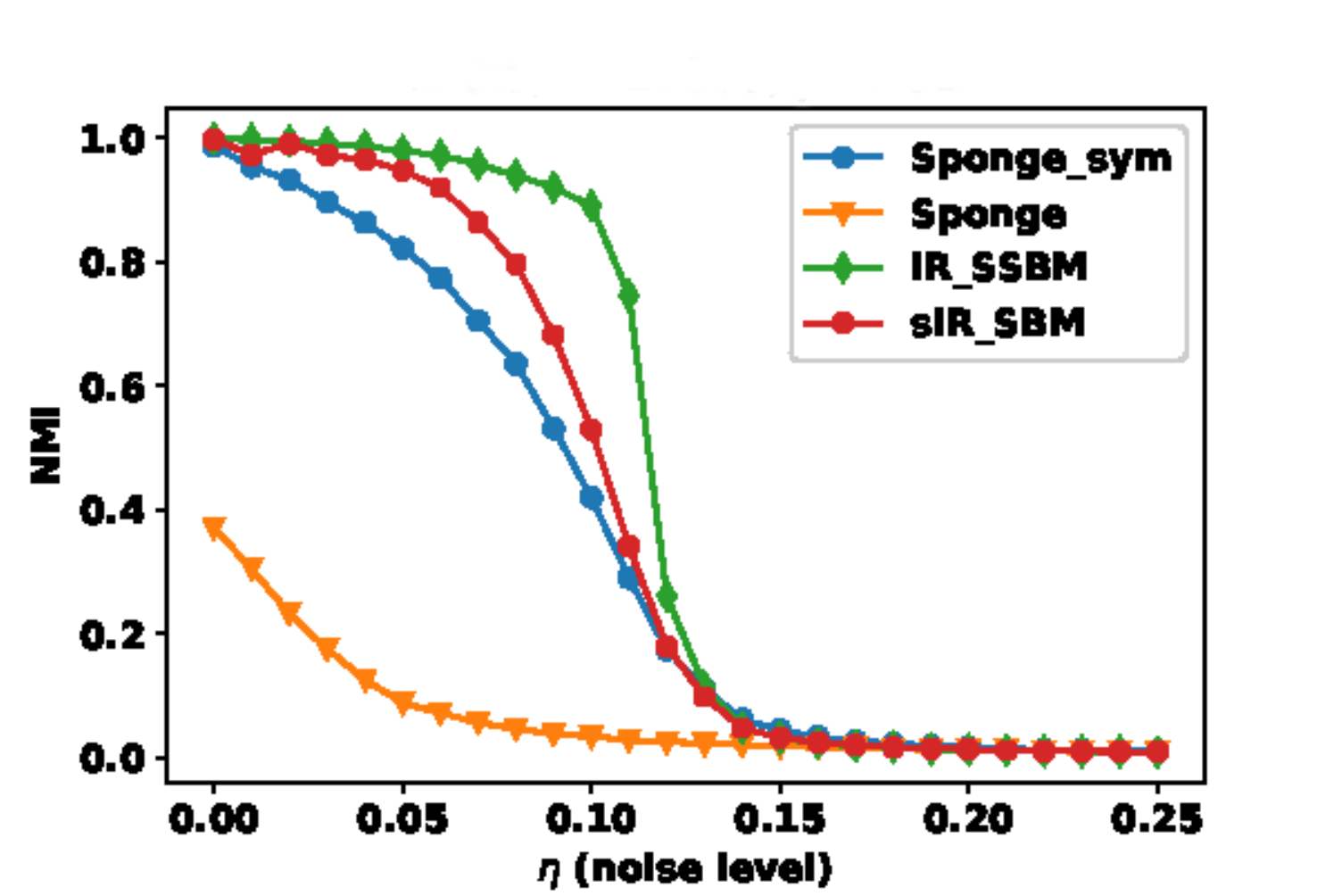}
    \caption{{\small NMI versus $\eta$ (noise) under signed SBM, $K=20$, $n=10000$, $p=0.01$.} }
    \label{fig:sponge}
\end{figure}

A graph is generated from the Signed SBM as follows. First we generate an Erdös-Renyi graph where each edge appears with probability $p$ and each edge takes the value $1$ if both extremities are in the same community and $-1$ otherwise. Then we flip the sign of each edge independently with probability $\eta \in [0,1/2)$. 
Our method \sircsbm\, can be directly applied to this setting, but we can also use the fact that the connectivity matrix $\Pi$ is assortative to design a more specialised algorithm \irssbm\, (see appendix) 
that  assigns a node to the community which maximizes its intra-connectivity estimated probability.
For initialization, we use \spongesym \,\citep{Cucuringu2019SPONGEAG} for clustering signed graphs. Figure \ref{fig:sponge} shows that $20$ refinement steps improve the clustering. 

\subsection{Australia Rainfall Dataset} \label{subsec:aus_dataset}
We consider the time series data of historical rainfalls in locations throughout Australia, this was also studied in  \cite{Cucuringu2019SPONGEAG}. Edge weights are obtained from the pairwise Pearson correlation, leading to a complete signed graph on $n = 306$ nodes. We use the longitude and latitude as covariates $X$, and \sponge\, \citep{Cucuringu2019SPONGEAG} to obtain an initial partition for  \sircsbm\, and \irsbm\, (the version of \sircsbm\, without covariates). We exclude $\ircsbm$ here due to its relative instability on this dataset  (see appendix). This shows that in some situations it can be better to use \sircsbm\, rather than \ircsbm. 
\begin{figure}
    \centering
    \includegraphics[scale=0.22]{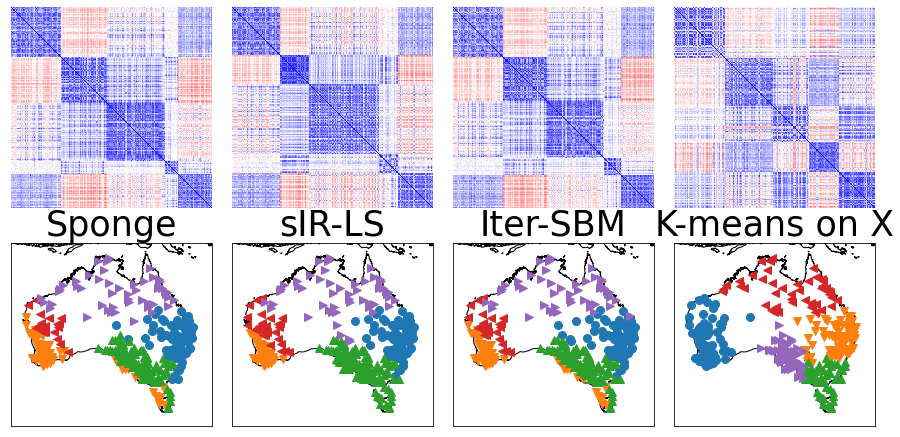}
    \caption{{\small  Sorted adjacency matrices 
    and maps for Australian rainfall dataset ($K=5$).}}
    \label{fig:aus}
\end{figure}
Figure \ref{fig:aus} illustrates the clustering obtained with \sponge\, (using only the graph), \sircsbm\, (integrating the covariates), \irsbm\, (refinement without covariates), and \kmeans\, applied on the covariates. The use of covariates in the refinement steps reinforces the geographical structure (orange points in the bottom right part of the map disappeared), increases the size of smallest cluster (the violet cluster on the three first maps), and strengthens the original clustering as seen in the sorted adjacency matrix, whereas \irsbm\, ignores the geography and \kmeans\, ignores the graph structure. Results for other choices of $K$ are in the appendix.
\section{Future work}
We only analyzed \sircsbm\, and \ircssbm\, to reduce  technicalities but we believe that the framework can be extended to analyze \ircsbm. The principle we used to design our algorithm could also be applied to obtain a clustering method for bipartite graphs or multilayer networks. Another direction of research is to extend our method to more general graph models integrating common properties of real-life networks such as degree heterogeneity, mixed membership, presence of outliers and missing values. 

\newpage
\typeout{}
\paragraph{Acknowledgement} G.B. would like to thank Pierre Latouche for fruitful discussions on clustering problems on graphs with side information.
\bibliography{biblio}

\begin{thebibliography}{39}
\providecommand{\natexlab}[1]{#1}
\providecommand{\url}[1]{\texttt{#1}}
\expandafter\ifx\csname urlstyle\endcsname\relax
  \providecommand{\doi}[1]{doi: #1}\else
  \providecommand{\doi}{doi: \begingroup \urlstyle{rm}\Url}\fi

\bibitem[Abbe(2018)]{AbbeSBM}
Abbe, E.
\newblock Community detection and stochastic block models.
\newblock \emph{Foundations and Trends® in Communications and Information
  Theory}, 14\penalty0 (1-2):\penalty0 1--162, 2018.

\bibitem[Abbe et~al.(2020)Abbe, Fan, and Wang]{abbe2020ellp}
Abbe, E., Fan, J., and Wang, K.
\newblock An $\ell_p$ theory of pca and spectral clustering, 2020.

\bibitem[Ahn et~al.(2018)Ahn, Lee, Cha, and Suh]{Ahn2018BinaryRE}
Ahn, K., Lee, K., Cha, H., and Suh, C.
\newblock Binary rating estimation with graph side information.
\newblock In \emph{NeurIPS}, 2018.

\bibitem[Bandeira \& van Handel(2016)Bandeira and van Handel]{bandeira2016}
Bandeira, A.~S. and van Handel, R.
\newblock Sharp nonasymptotic bounds on the norm of random matrices with
  independent entries.
\newblock \emph{Ann. Probab.}, 44\penalty0 (4):\penalty0 2479--2506, 07 2016.

\bibitem[Binkiewicz et~al.(2017)Binkiewicz, Vogelstein, and
  Rohe]{Binkiewicz2017CovariateassistedSC}
Binkiewicz, N., Vogelstein, J., and Rohe, K.
\newblock Covariate-assisted spectral clustering.
\newblock \emph{Biometrika}, 104:\penalty0 361 -- 377, 2017.

\bibitem[Boumal(2016)]{boumal2016}
Boumal, N.
\newblock Nonconvex phase synchronization.
\newblock \emph{SIAM Journal on Optimization}, 26, 01 2016.

\bibitem[Celeux \& Govaert(1992)Celeux and Govaert]{Celeux1992ACE}
Celeux, G. and Govaert, G.
\newblock A classification em algorithm for clustering and two stochastic
  versions.
\newblock \emph{Computational Statistics \& Data Analysis}, 14:\penalty0
  315--332, 1992.

\bibitem[Chen \& Zhang(2021)Chen and Zhang]{Chen2021OptimalCI}
Chen, X. and Zhang, A.
\newblock Optimal clustering in anisotropic gaussian mixture models.
\newblock \emph{ArXiv}, 2021.

\bibitem[Chen \& Candès(2016)Chen and Candès]{chen2016_alignment}
Chen, Y. and Candès, E.
\newblock The projected power method: An efficient algorithm for joint
  alignment from pairwise differences.
\newblock \emph{Communications on Pure and Applied Mathematics}, 71, 09 2016.

\bibitem[Chi et~al.(2019{\natexlab{a}})Chi, Lu, and Chen]{NCOpt2019}
Chi, Y., Lu, Y., and Chen, Y.
\newblock Nonconvex optimization meets low-rank matrix factorization: An
  overview.
\newblock \emph{IEEE Transactions on Signal Processing}, PP:\penalty0 1--1, 08
  2019{\natexlab{a}}.
\newblock \doi{10.1109/TSP.2019.2937282}.

\bibitem[Chi et~al.(2019{\natexlab{b}})Chi, Lu, and Chen]{chi2019}
Chi, Y., Lu, Y., and Chen, Y.
\newblock Nonconvex optimization meets low-rank matrix factorization: An
  overview.
\newblock \emph{IEEE Transactions on Signal Processing}, PP:\penalty0 1--1, 08
  2019{\natexlab{b}}.
\newblock \doi{10.1109/TSP.2019.2937282}.

\bibitem[Contisciani et~al.(2020)Contisciani, Power, and
  De~Bacco]{contisciani2020}
Contisciani, M., Power, E., and De~Bacco, C.
\newblock Community detection with node attributes in multilayer networks.
\newblock \emph{Scientific reports}, 10:\penalty0 15736, 09 2020.
\newblock \doi{10.1038/s41598-020-72626-y}.

\bibitem[Cucuringu et~al.(2019)Cucuringu, Davies, Glielmo, and
  Tyagi]{Cucuringu2019SPONGEAG}
Cucuringu, M., Davies, P., Glielmo, A., and Tyagi, H.
\newblock Sponge: A generalized eigenproblem for clustering signed networks.
\newblock In \emph{AISTATS}, 2019.

\bibitem[Deshpande et~al.(2018)Deshpande, Montanari, Mossel, and
  Sen]{Deshpande2018ContextualSB}
Deshpande, Y., Montanari, A., Mossel, E., and Sen, S.
\newblock Contextual stochastic block models.
\newblock In \emph{NeurIPS}, 2018.

\bibitem[Emmons \& Mucha(2019)Emmons and Mucha]{emmons2019}
Emmons, S. and Mucha, P.~J.
\newblock Map equation with metadata: Varying the role of attributes in
  community detection.
\newblock \emph{Phys. Rev. E}, 100:\penalty0 022301, Aug 2019.

\bibitem[Fajardo-Fontiveros et~al.(2022)Fajardo-Fontiveros, Guimerà, and
  Sales-Pardo]{fontiveros2022}
Fajardo-Fontiveros, O., Guimerà, R., and Sales-Pardo, M.
\newblock Node metadata can produce predictability crossovers in network
  inference problems.
\newblock \emph{Physical Review X}, 12, 01 2022.

\bibitem[Gao \& Zhang(2019)Gao and Zhang]{Gao2019IterativeAF}
Gao, C. and Zhang, A.
\newblock Iterative algorithm for discrete structure recovery.
\newblock \emph{arXiv: Statistics Theory}, 2019.

\bibitem[Gao et~al.(2016)Gao, Ma, Zhang, and Zhou]{DCBM2016}
Gao, C., Ma, Z., Zhang, A., and Zhou, H.
\newblock Community detection in degree-corrected block models.
\newblock \emph{Annals of Statistics}, 46, 07 2016.

\bibitem[Han et~al.(2020)Han, Luo, Wang, and Zhang]{Han2020ExactCI}
Han, R., Luo, Y., Wang, M., and Zhang, A.
\newblock Exact clustering in tensor block model: Statistical optimality and
  computational limit.
\newblock \emph{ArXiv}, 2020.

\bibitem[Holland et~al.(1983)Holland, Laskey, and Leinhardt]{HOLLAND1983109}
Holland, P.~W., Laskey, K.~B., and Leinhardt, S.
\newblock Stochastic blockmodels: First steps.
\newblock \emph{Social Networks}, 5\penalty0 (2):\penalty0 109 -- 137, 1983.

\bibitem[Hric et~al.(2016)Hric, Peixoto, and Fortunato]{hric2016}
Hric, D., Peixoto, T., and Fortunato, S.
\newblock Network structure, metadata, and the prediction of missing nodes and
  annotations.
\newblock \emph{Physical Review X}, 6, 09 2016.
\newblock \doi{10.1103/PhysRevX.6.031038}.

\bibitem[Jin et~al.(2021)Jin, Ke, and Luo]{Jin2021OptimalAO}
Jin, J., Ke, Z.~T., and Luo, S.
\newblock Optimal adaptivity of signed-polygon statistics for network testing.
\newblock \emph{The Annals of Statistics}, 2021.

\bibitem[L{\'e}ger(2016)]{r2016Blockmodels}
L{\'e}ger, J.-B.
\newblock Blockmodels: A r-package for estimating in latent block model and
  stochastic block model, with various probability functions, with or without
  covariates.
\newblock \emph{arXiv: Computation}, 2016.

\bibitem[Lu \& Sen(2020)Lu and Sen]{lu2020contextual}
Lu, C. and Sen, S.
\newblock Contextual stochastic block model: Sharp thresholds and contiguity,
  2020.

\bibitem[Lu \& Zhou(2016)Lu and Zhou]{Lu2016StatisticalAC}
Lu, Y. and Zhou, H.~H.
\newblock Statistical and computational guarantees of lloyd's algorithm and its
  variants.
\newblock \emph{ArXiv}, 2016.

\bibitem[Ma \& Nandy(2021)Ma and Nandy]{ma2021community}
Ma, Z. and Nandy, S.
\newblock Community detection with contextual multilayer networks, 2021.

\bibitem[Mele et~al.(2019)Mele, Hao, Cape, and Priebe]{Mele_2019}
Mele, A., Hao, L., Cape, J., and Priebe, C.
\newblock Spectral inference for large stochastic blockmodels with nodal
  covariates.
\newblock \emph{SSRN Electronic Journal}, 2019.

\bibitem[Mouselimis(2021)]{clusterr}
Mouselimis, L.
\newblock \emph{{ClusterR}: Gaussian Mixture Models, K-Means,
  Mini-Batch-Kmeans, K-Medoids and Affinity Propagation Clustering}, 2021.

\bibitem[Ndaoud et~al.(2019)Ndaoud, Sigalla, and
  Tsybakov]{Ndaoud2019ImprovedCA}
Ndaoud, M., Sigalla, S., and Tsybakov, A.
\newblock Improved clustering algorithms for the bipartite stochastic block
  model.
\newblock \emph{arXiv: Statistics Theory}, 2019.

\bibitem[Newman \& Clauset(2015)Newman and Clauset]{newman2016}
Newman, M. and Clauset, A.
\newblock Structure and inference in annotated networks.
\newblock \emph{Nature Communications}, 7, 07 2015.

\bibitem[Onaran \& Villar(2017)Onaran and Villar]{gpmmatching}
Onaran, E. and Villar, S.
\newblock Projected power iteration for network alignment.
\newblock pp.\ ~45, 08 2017.
\newblock \doi{10.1117/12.2275366}.

\bibitem[Stanley et~al.(2019)Stanley, Bonacci, Kwitt, Niethammer, and
  Mucha]{stanley2019}
Stanley, N., Bonacci, T., Kwitt, R., Niethammer, M., and Mucha, P.
\newblock Stochastic block models with multiple continuous attributes.
\newblock \emph{Applied Network Science}, 4, 08 2019.
\newblock \doi{10.1007/s41109-019-0170-z}.

\bibitem[Wang et~al.(2021{\natexlab{a}})Wang, Liu, Zhou, and
  So]{Wang2021OptimalNE}
Wang, P., Liu, H., Zhou, Z., and So, A. M.-C.
\newblock Optimal non-convex exact recovery in stochastic block model via
  projected power method.
\newblock In \emph{ICML}, 2021{\natexlab{a}}.

\bibitem[Wang et~al.(2021{\natexlab{b}})Wang, Liu, Zhou, and So]{wang21pm}
Wang, P., Liu, H., Zhou, Z., and So, A. M.-C.
\newblock Optimal non-convex exact recovery in stochastic block model via
  projected power method.
\newblock In \emph{Proceedings of the 38th International Conference on Machine
  Learning}, volume 139 of \emph{Proceedings of Machine Learning Research},
  pp.\  10828--10838. PMLR, 18--24 Jul 2021{\natexlab{b}}.

\bibitem[Weng \& Feng(2016)Weng and Feng]{Weng2016CommunityDW}
Weng, H. and Feng, Y.
\newblock Community detection with nodal information.
\newblock \emph{arXiv: Methodology}, 2016.

\bibitem[Yan \& Sarkar(2020)Yan and Sarkar]{CovReg}
Yan, B. and Sarkar, P.
\newblock {Covariate Regularized Community Detection in Sparse Graphs}.
\newblock \emph{Journal of the American Statistical Association}, 0\penalty0
  (0):\penalty0 1--12, 2020.

\bibitem[Yang et~al.(2013)Yang, McAuley, and Leskovec]{Yang2013CommunityDI}
Yang, J., McAuley, J., and Leskovec, J.
\newblock Community detection in networks with node attributes.
\newblock \emph{2013 IEEE 13th International Conference on Data Mining}, pp.\
  1151--1156, 2013.

\bibitem[Zhang \& Zhou(2016)Zhang and Zhou]{MiniMaxZhang2016}
Zhang, A.~Y. and Zhou, H.~H.
\newblock {Minimax Rates of Community Detection in Stochastic Block Models}.
\newblock \emph{The Annals of Statistics}, 44\penalty0 (5):\penalty0
  2252--2280, sep 2016.
\newblock ISSN 00905364.

\bibitem[Zhang et~al.(2015)Zhang, Levina, and Zhu]{Zhang2015CommunityDI}
Zhang, Y., Levina, E., and Zhu, J.
\newblock Community detection in networks with node features.
\newblock \emph{ArXiv}, 2015.

\end{thebibliography}
\bibliographystyle{style_icml/icml2022}
\newpage

\appendix
\onecolumn
\begin{center}
\Large \textbf{Supplementary Material}
\end{center}

The proof of Corollary \ref{cor:contraction} is presented in Section \ref{app:coro} and the proof of Lemma \ref{lem:ideal_error} is presented in Section \ref{app:ideal_error}. Theorem \ref{thm:IRcv} is proved in Section \ref{app:IRcv} and Theorem \ref{thm:minimax} is proved in Section \ref{app:minimax}. The technical lemmas used in the proofs are gathered in Section \ref{app:lemmas}. Finally, Section \ref{app:xp} contains results for additional experiments.

\section{Proof of Corollary \ref{cor:contraction}}\label{app:coro}
We will prove a more precise version of Corollary \ref{cor:contraction} stated below.
\begin{corollary} Assume that the assumptions of Theorem $1$ hold. Denoting $t^*= \lceil \frac{\log (1/\delta)}{\log (8/7)} \rceil$, we have for all $t\geq t^*$ that $\tau^{(t)} = \tau$, $\delta^{(t)} = \delta$ and  \[ \lt \leq \frac{8}{7}\xi(\delta)+ \frac{3}{28} \tau^{(0)}\left( \frac{1}{8}\right) ^{t-t^*}\] 
\end{corollary}

\begin{proof}
For convenience, denote $\tau = \tau^{(0)} \delta$. Let $t^*$ be the smallest integer such that $\tau \geq ( \frac{7}{8})^t\tau^{(0)}$; clearly,  $t^*=\lceil \frac{\log (1/\delta)}{\log (8/7)} \rceil$. Then for $t \geq t^*$, we have $\tau^{(t)} = \tau = \tau^{(0)} \delta$, and hence (from the definition of $\delta^{(t)}$), $\delta^{(t)} = \delta$. Therefore Theorem \ref{thm:gao_ext} implies that for $t\geq t^*$,
\begin{align*}
    \lt & \leq \xi(\delta^{(t)} ) + \frac{1}{8} \xi(\delta^{(t-1)}) + \ldots + \left(\frac{1}{8}\right)^{t-t^*}\xi (\delta^{(t^*)})+\left[\left(\frac{1}{8}\right)^{t-t^*+1}\xi (\delta^{(t^*-1)})+ \ldots + \left(\frac{1}{8}\right)^{t-1}\xi (\delta^{(1)}) \right] \\
    &\leq \frac{8}{7}\xi(\delta) + \left[\left(\frac{1}{8}\right)^{t-t^*+1}\xi (\delta^{(t^*-1)})+ \ldots + \left(\frac{1}{8}\right)^{t-1}\xi (\delta^{(1)}) \right] \tag{since $\delta^{(t)}=\delta$ for $t\geq t^*$}\\
    &\leq \frac{8}{7}\xi(\delta) +\frac{3}{4}\tau^{(0)}\left[\left(\frac{1}{8}\right)^{t-1}+ \ldots + \left(\frac{1}{8}\right)^{t-t^*+1} \right] \tag{since for all $t$, $\xi(\delta^{(t)})\leq \frac{3}{4}\tau^{(0)}$ by Condition \ref{cond:ideal}}\\
    &\leq \frac{8}{7}\xi(\delta) + \frac{3}{28}\tau^{(0)}\left(\frac{1}{8}\right)^{t-t^*}.
\end{align*}
\end{proof}

\section{Proof of Lemma \ref{lem:ideal_error}}\label{app:ideal_error}

Let $\delta, \bar{\delta}>0$ . The ideal oracle error term can be upper bounded as follows 
\begin{align*}
    \xi(\delta) &\leq \underbrace{ \sum_{i=1}^n \sum_{b \in [K]\backslash z_i} \Delta^2(z_i,b)\indic_{\lbrace \langle \epsilon_i, \mu_{z_i}-\mu_b\rangle+ \lambda \langle E_{i:}W, \Pi_{z_i:}-\Pi_{b:}\rangle \leq \frac{-(1-\delta-\Bar{\delta})\Delta^2(z_i,b)}{2}\rbrace } }_{M_1}\\
    &+\underbrace{\sum_{i=1}^n \sum_{b \in [K]\backslash z_i} \Delta^2(z_i,b)\indic_{\lbrace \langle \epsilon_i, \Tilde{\mu}_{z_i}-\mu_{z_i}\rangle+ \lambda \langle E_{i:}W, \Tilde{\Pi}_{z_i:}-\Pi_{z_i:}\rangle \leq \frac{-\Bar{\delta}\Delta^2(z_i,b)}{4}\rbrace } }_{M_2}\\
    &+\underbrace{\sum_{i=1}^n \sum_{b \in [K]\backslash z_i} \Delta^2(z_i,b)\indic_{\lbrace \langle -\epsilon_i, \Tilde{\mu}_b-\mu_b\rangle- \lambda \langle E_{i:}W, \Tilde{\Pi}_{b:}-\Pi_{b:}\rangle \leq \frac{-\Bar{\delta}\Delta^2(z_i,b)}{4}\rbrace } }_{M_3}.
\end{align*}
We will first obtain upper bounds for each $\expec(M_i)$, $i=1\ldots 3$. In particular we will show that the dominant term is $\expec(M_1)$. Then, we will use Markov inequality to control $\xi(\delta)$ with high probability.

\paragraph{Upper bound of $\expec(M_1)$.}
Let us denote for any given $i\in [n]$ and $b\in [K]\setminus z_i$ the event \[\Omega_1 = \left\lbrace \langle \epsilon_i, \mu_{z_i}-\mu_b\rangle+ \lambda \langle E_{i:}W, \Pi_{z_i:}-\Pi_{b:}\rangle \leq \frac{-(1-\delta-\Bar{\delta})\Delta^2(z_i,b)}{2} \right\rbrace. \]
By using an analogous argument as the one presented in Lemma \ref{lem:ideal_error_ssbm} we obtain \[ \prob (\Omega_1) \leq \exp (-(1+o(1))\Tilde{\Delta}^2 ). \] 

Thus by taking $\delta=\bar{\delta}$ going to zero as $n \to \infty$, we obtain  
\begin{align*}
     \expec (M_1)& \leq  \sum_{i=1}^n\sum_{b\in[K]\backslash z_i}\Delta^2(z_i,b)\exp(-(1+o(1))\Tilde{\Delta}^2)\\
     &\leq nK\exp(-(1+o(1))\Tilde{\Delta}^2)\\
     &\leq n\exp(-(1+o(1))\Tilde{\Delta}^2).
\end{align*}  
In the second line we used the fact that $\Delta^2(z_i,b) =\Delta_{min} \lesssim \tilde{\Delta}^2$ for all $z_i\neq b$ and $\tilde{\Delta}\asymp \sqrt{\log n/K} \to \infty$. In the third line we used the assumption  $\Tilde{\Delta}^2/ \log(K)\to \infty$.

\paragraph{Upper bound of $\expec(M_2)$.}
Let us denote for any given $i\in [n]$ and $b\in [K]\setminus z_i$ the events 
\[ \Omega_2= \left\lbrace \langle \epsilon_i, \Tilde{\mu}_{z_i}-\mu_{z_i}\rangle+ \lambda \langle E_{i:}W, \Tilde{\Pi}_{z_i:}-\Pi_{z_i:}\rangle \leq \frac{-\Bar{\delta}\Delta^2(z_i,b)}{4} \right\rbrace, \]
 \[ \Omega_2'= \left\lbrace \langle \epsilon_i, \Tilde{\mu}_{z_i}-\mu_{z_i}\rangle\leq \frac{-\Bar{\delta}\Delta^2(z_i,b)}{8} \right\rbrace, \]
 and \[ \Omega_2''= \left\lbrace \lambda \langle E_{i:}W, \Tilde{\Pi}_{z_i:}-\Pi_{z_i:}\rangle\leq \frac{-\Bar{\delta}\Delta^2(z_i,b)}{8} \right\rbrace.\]
Clearly $\prob(\Omega_2)\leq \prob(\Omega_2')+\prob(\Omega_2'')$ by a union bound argument.
 
 Let us first upper bound $\prob(\Omega_2')$. Recall that $n_k=n/K$ under the CSSBM by assumption, let us also define $n_{min} := \min_{k} n_k$. We keep this general notation because it is shorter and indicates how the proof can be generalized to the unbalanced setting. By definition $\tilde{\mu}_{z_i}-\mu_{z_i}=\sum_{j\in \calC_{z_i}} \frac{\epsilon_j}{n_{z_i}}$, hence \[ \langle \epsilon_i, \Tilde{\mu}_{z_i}-\mu_{z_i}\rangle = \frac{||\epsilon_i||^2+\epsilon_i^\top \sum_{\substack{j \in \calC_{z_i}\\j\neq i}}\epsilon_j}{n_{z_i}}.\] 
 This last quantity is lower bounded by $ \epsilon_i^\top \eta_i$ where $\eta_i =\frac{ \sum_{j \in \calC_{z_i}, \, j\neq i}\epsilon_j}{n_{z_i}} $. In particular $\epsilon_i$ and $\eta_i$ are independent and their entries are also independent. Moreover $\eta_i$ is a centered gaussian random variables with independent entries such that $\Var((\eta_i)_k) \leq 1/n_k$. So by Bernstein inequality, it holds for all $x>0$ that 
 \[ \prob\left(|| \eta_i||^2 \geq \frac{1}{n_k}(K+2\sqrt{Kx}+2x) \right) \leq \exp(-x) \]
which in turn implies 
\begin{align*}
    \prob\left(\epsilon_i^\top\eta_i \leq -\frac{\bar{\delta} \Delta^2_{min}}{8} \right) &\leq  \prob\left(\epsilon_i^\top\eta_i \leq -\frac{\bar{\delta} \Delta^2_{min}}{8} \;\middle|\; || \eta_i||^2 \leq \frac{1}{n_k}(K+2\sqrt{Kx}+2x)\right)\\
    &+\prob\left(|| \eta_i||^2 \geq \frac{1}{n_k}(K+2\sqrt{Kx}+2x)\right)\\
    & \leq \exp\left(-c\frac{n_k(\Bar{\delta}\Delta_{min}^2)^2}{K+2\sqrt{Kx}+2x}\right)+\exp(-x).
\end{align*}
Setting $x= \sqrt{n_k} \bar{\delta}\Delta^2_{min}$ we obtain $\prob(\Omega_2')\leq 2 \exp(-C\bar{\delta} \sqrt{n_k}\Delta^2_{min})$. Since $\bar{\delta} \to 0$ (as $n \rightarrow \infty$) at a suitably slow rate, we have $\bar{\delta} \sqrt{n/K}\to +\infty$.  Consequently, 
\[ \prob(\Omega_2')= o(\exp (-(C+o(1))\Delta_{min}^2) = o(\exp(-(1+o(1))\Tilde{\Delta}^2)).\]

Let us now bound $\prob(\Omega_2'')$. Since $\tilde{\Pi}_{kk'}- \Pi_{kk'}= \sum_{i \in \calC_k, \, j\in \calC_{k'}}\frac{E_{ij}}{n_kn_{k'}} = W_{:k}^\top E W_{:k'}$, we obtain the decomposition 
\[ \lambda \langle E_{i:}W, \Tilde{\Pi}_{z_i:}-\Pi_{z_i:}\rangle =\lambda \langle E_{i:}W, W_{z_i:}^\top E^{(i)}W \rangle + \lambda\langle E_{i:}W, W_{z_i:}^\top E^{(-i)}W \rangle \] 
where $E^{(i)}$ is obtained from $E$ by only keeping the $i$th row and column, and $E_{-i}$ is the matrix obtained from $E$ by replacing the $i$th row and column by zero. In particular, $E^{(-i)}$ is independent from $E_{i:}$. The second term can be controlled by using the same techniques as before. Indeed, the entries of $W_{z_i:}^\top E^{(-i)}W $ are independent and $\Var(W_{z_i:}^\top E^{(-i)}W_{:k})\leq C\frac{p_{max}}{n_{min}^2}$ for all $k$. Denoting  $\eta_{ik}' =W_{z_i:}^\top E^{(-i)}W_{:k}$ and $\eta_i'=(\eta_{ik}')_{k\in [K]}$, 
this implies  
\begin{align*}
    \prob \left(  \lambda \langle E_{i:}W,\eta_i' \rangle  \leq -\frac{\bar{\delta}\Delta^2_{min}}{16} \right) &\leq  \prob \left( \lambda \langle E_{i:}W, \eta_i'\rangle \leq -\frac{\bar{\delta}\Delta_{min}^2}{16} \;\middle|\; \forall k,\,| \eta_{ik}'|^2 \leq C\frac{p_{max}}{n_k^2}(K+x)\right) \\
    &+ \prob\left(\exists k,\,| \eta_{ik}'|^2 \geq C\frac{p_{max}}{n_{min}^2}(K+x)  \right)\\
    & \leq K\exp\left(-C \frac{n_{min}^3(\bar{\delta}\Delta_{min}^2)^2}{p_{max}\lambda^2(K+x)}\right) +K\exp(-x).
\end{align*} 
Here, we used a union bound argument for the first inequality. The second inequality uses Lemma \ref{lem:chern_bin} -- which provides a concentration bound for binomial random variables -- along with the fact $\lambda \asymp \frac{n}{Kp_{max}}$.

Setting $x = C\sqrt{n_kp_{max}}\bar{\delta} \Delta_{min}^2$ we obtain 
\begin{align*}
    \prob \left(  \lambda \langle E_{i:}W,\eta_i' \rangle  \leq -\frac{\bar{\delta}\Delta^2_{min}}{16} \right)  \leq 2\exp(-c\sqrt{n_kp_{max}}\bar{\delta} \Delta_{min}^2) 
    = o(\exp(-\Tilde{\Delta}^2))
\end{align*} 
since  $\bar{\delta}$ can be chosen such that $\sqrt{n_kp_{max}}\bar{\delta}\to + \infty$ (because by assumption $np_{max}\asymp \log n\gg K^4$) and $\Delta_{min}\asymp \tilde{\Delta}$.

It remains to control $\langle E_{i:}W, W_{z_i:}^\top E^{(i)}W \rangle $. Using the fact 
\begin{equation*}
    (W_{z_i:}^\top E^{(i)}W)_k=
    \begin{cases}
    \sum_{j'\in \calC_k}\frac{E_{ij'}}{n_{z_i}n_k} & \text{if }k \neq z_i\\
    2\sum_{j'\in \calC_k}\frac{E_{ij'}}{n_{z_i}^2} & \text{if } k=z_i
    \end{cases}
\end{equation*}
we have 
\begin{align*}
    \langle E_{i:}W, W_{z_i:}^\top E^{(i)}W \rangle &= \sum_{k\neq z_i} \sum_{\substack{j \in \calC_k\\j' \in \calC_k }}\frac{E_{ij}}{n_k}\frac{E_{ij'}}{n_{z_i}n_k}+ 2n_{z_i}\left(\sum_{j\in \calC_{z_i}}\frac{E_{ij'}}{n_{z_i}^2}\right)^2\\
    &= \frac{1}{n_{z_i}}\left( \sum_{j\in \calC_k}\frac{E_{ij}}{n_k}\right)^2+ 2n_{z_i} \left(\sum_{j\in \calC_{z_i}}\frac{E_{ij'}}{n_{z_i}^2}\right)^2 \\
    & \geq 0.
\end{align*}  
Consequently, $\prob\left( \Omega_2'' \right)$ can be bounded as 
\[ \prob\left( \Omega_2'' \right) \leq  \prob \left(  \lambda \langle E_{i:}W,\eta_i' \rangle  \leq -\frac{\bar{\delta}\Delta_{min}^2}{16} \right)= o(\exp(-\Tilde{\Delta}^2)).\] 
%

\paragraph{Upper bound of $\expec(M_3)$.}  Let us denote for any given $i\in [n]$ and $b\in [K]\setminus z_i$ the event \[ \Omega_3 = \left\lbrace \langle -\epsilon_i, \Tilde{\mu}_b-\mu_b\rangle- \lambda \langle E_{i:}W, \Tilde{\Pi}_{b:}-\Pi_{b:}\rangle \leq \frac{-\Bar{\delta}\Delta^2(z_i,b)}{4} \right\rbrace .\]
First observe that \[ \langle \epsilon_i, \Tilde{\mu}_{b}-\mu_{b}\rangle =  \frac{\epsilon_i^\top \sum_{j\in \calC_b}\epsilon_j}{n_{z_i}},\] therefore this term can be handled in the same way as before.
Moreover, we have \[ \lambda \langle E_{i:}W, \Tilde{\Pi}_{b:}-\Pi_{b:}\rangle =\lambda \langle E_{i:}W, W_{b:}^\top E^{(i)}W \rangle + \lambda\langle E_{i:}W, W_{b:}^\top E^{(-i)}W \rangle . \]
The second term can be handled in the same way as before by using a conditioning argument.
Now observe that \[ \langle E_{i:}W, W_{b:}^\top E^{(i)}W \rangle = \frac{1}{n_{z_i}^2n_b}\left(\sum_{j\in\calC_{b}}E_{ij} \right) \left(\sum_{j'\in\calC_{z_i}}E_{ij'} \right) \]
 where $\sum_{j\in\calC_{b}}E_{ij}$ and $\sum_{j'\in\calC_{z_i}}E_{ij'}$ are independent subgaussian random variables. Thus this term can also be controlled by using the same conditioning argument as before.
 
\paragraph{Conclusion.}
The previously obtained upper bounds imply 
\[ \expec(\xi(\delta)) \leq 3 \expec(M_1) \leq n\exp(-(1+o(1))\Tilde{\Delta}^2). \]
Finally, by Markov inequality, we obtain \[ \prob (\xi(\delta)\geq \exp(\tilde{\Delta})\expec\xi(\delta))  \leq \exp(-\tilde{\Delta}).\]
But since \[ \exp(\tilde{\Delta})\expec\xi(\delta) \leq n\exp(-(1+o(1))\tilde{\Delta}^2)\] we obtain that with probability at least $1-\exp(-\tilde{\Delta})$ \[ \xi(\delta) \leq n\exp(-(1+o(1))\Tilde{\Delta}^2).\]

\section{Proof of Theorem \ref{thm:IRcv}}\label{app:IRcv}
The general proof strategy has been presented in Section \ref{sec:gen_fr}. In Section \ref{app:error-dec} we will make the error decomposition explicit.  Then, we will control the different error terms in Sections \ref{app:f_error}, \ref{app:g_error} and \ref{app:h_error}. Finally, we will conclude by applying Theorem \ref{thm:gao_ext} in Section \ref{app:thm2_cl}.

\subsection{ Error decomposition for the one-step analysis of \ircssbm}\label{app:error-dec}
We will assume without lost of generality that $\sigma=1$ to simplify the exposition. Let $i \in [n]$ and $a\in [K]$ be such that\footnote{Depending on the context we will interchangeably use the notation $z_i$ and $a$.} $z_i=a$, and let 
\[ \lambda^{(t)}=  \frac{n}{K(p^{(t)}-q^{(t)})} \log \left(\frac{p^{(t)}(1-q^{(t)})}{q^{(t)}(1-p^{(t)})}\right)\] 
denote the scalar corresponding to the diagonal entry of the inverse covariance matrix $\Sigma_k^{(t)}$. Similarly, let us denote \[  \lambda=  \frac{n}{K(p-q)} \log \left(\frac{p(1-q)}{q(1-p)}\right).\]  Given the current estimator of the partition $Z^{(t)}$,  node $i$ will be incorrectly estimated after one refinement step if
    \[ a \neq \arg \min_k  ||X_i-\mu_k ^{(t)}||^2+ \hat{\lambda}^{(t)}||A_{i:}W^{(t)}-\Pi^{(t)}_{k:} ||^2 \]
 or equivalently, if there exists $b\in [K]\backslash a$ such that
 \begin{equation*}
 ||X_i-\mu_b ^{(t)}||^2+ \hat{\lambda}^{(t)}||A_{i:}W^{(t)}-\Pi^{(t)}_{b:} ||^2 \leq ||X_i-\mu_a ^{(t)}||^2+\hat{\lambda}^{(t)}||A_{i:}W^{(t)}-\Pi^{(t)}_{a:} ||^2.
 \end{equation*} 
The above inequality is equivalent to 
\[
    \langle \epsilon_i, \Tilde{\mu}_a-\Tilde{\mu}_b\rangle+ \lambda \langle E_{i:}W, \Tilde{\Pi}_{a:}-\Tilde{\Pi}_{b:}\rangle \leq \frac{-\Delta^2(a,b)}{2} +F_{ib}^{(t)}+G_{ib}^{(t)}+H_{ib}^{(t)}
\]
where
\[\Delta^2(a,b)= ||\mu_a-\mu_b ||^2+ \lambda ||\Pi_{a:}-\Pi_{b:} ||^2, \ \Tilde{\mu}_k = X^\top W_{:a}, \ \text{and} \ \Tilde{\Pi}_{k:}=W_{k:}^\top AW \] 
for all $k\in [K]$. Furthermore, the terms $F_{ib}^{(t)}, G_{ib}^{(t)}$ and $H_{ib}^{(t)}$ are given by 
\begin{align*}
    F_{ib}^{(t)} &= \langle \epsilon_i, (\Tilde{\mu}_a-\mu_a^{(t)})-(\Tilde{\mu}_b-\mu_b^{(t)})\rangle + \lambda^{(t)}\langle E_{i:}W^{(t)}, (\Tilde{\Pi}_{a:}-\Pi_{a:}^{(t)})-(\Tilde{\Pi}_{b:}-\Pi_{b:}^{(t)})\rangle \\
    &+\lambda^{(t)}\langle E_{i:}(W-W^{(t)}),\Tilde{\Pi}_{a:}-\Tilde{\Pi}_{b:} \rangle + (\lambda - \lambda^{(t)})\langle E_{i:}W,\Tilde{\Pi}_{a:}-\Tilde{\Pi}_{b:}  \rangle, \\
        2G_{ib}^{(t)} &=(||\mu_a-\mu_a^{(t)} ||^2-||\mu_a-\tilde{\mu}_a ||^2)-(||\mu_a-\mu_b^{(t)} ||^2-||\mu_a-\tilde{\mu}_b ||^2)\\
    &+ \lambda^{(t)} (|| P_{i:}W^{(t)}-\Pi_{a:}^{(t)} ||^2-||P_{i:}W^{(t)}-W_{a:}^\top AW^{(t)} ||^2)\\
    &- \lambda^{(t)} (|| P_{i:}W^{(t)}-\Pi_{b:}^{(t)} ||^2-||P_{i:}W^{(t)}-W_{b:}^\top AW^{(t)} ||^2) \\
  \text{ and } \quad  2H_{ib}^{(t)} & =||\mu_a-\tilde{\mu}_a||^2-||\mu_a-\tilde{\mu}_b ||^2+ ||\mu_a-\mu_b||^2 \\
    &+ \lambda^{(t)} (||P_{i:}W^{(t)}-W_{a:}^\top AW^{(t)} ||^2 - ||P_{i:}W^{(t)}-W_{b:}^\top AW^{(t)} ||^2 + ||\Pi_{a:}-\Pi_{b:}||^2)\\
    &+ (\lambda-\lambda^{(t)})||\Pi_{a:}-\Pi_{b:}||^2.
\end{align*}
The main term in this decomposition is 
\[  \langle \epsilon_i, \Tilde{\mu}_a-\Tilde{\mu}_b\rangle+ \lambda \langle E_{i:}W, \Tilde{\Pi}_{a:}-\Tilde{\Pi}_{b:}\rangle \leq \frac{-\Delta^2(a,b)}{2} \] 
and corresponds to the error when the current estimation of the partition is the ground truth partition. It is controlled by Lemma \ref{lem:ideal_error}

The three error terms will be controlled in different ways. The error term $F_{ib}^{(t)}$ depends in a crucial way on $i$ and $t$, it will be controlled with a $l_2$-type norm (see Condition \ref{cond:f}). The square of the error terms $G_{ib}^{(t)}$ and $|H_{ib}^{(t)}|$ will be controlled uniformly (see Condition \ref{cond:h}).


\subsection{Bounding the error term $F_{ib}^{(t)}$}\label{app:f_error}
In this section we are going to show that Condition \ref{cond:f} is satisfied.
\begin{lemma}\label{lem:f} Under the assumptions of Theorem \ref{thm:sirls} (that are also satisfied by Theorem \ref{thm:IRcv}) we have w.h.p. that for all $z^{(t)}$ such that $\lt \leq \tau^{(0)}$,
\[\sum_{i=1}^n\max_{b\in [K]\backslash z_i} \frac{(F_{ib}^{(t)})^2}{\Delta^2(z_i,b)l(z,z^{(t)})} \leq \frac{\delta^2}{256}.\]
\end{lemma}

\begin{proof}

We need to upper-bound of \[ F=\max_{\lbrace z^{(t)}: l(z,z^{(t)})\leq \tau^{(0)} \rbrace}\sum_{i=1}^n\max_{b\in [K]\backslash z_i} \frac{(F_{ib}^{(t)})^2}{\Delta(z_i,b)^2l(z,z^{(t)})}.\]
To this end, we can decompose $F_{ib}^{(t)} = F_{ib}^{1,(t)}+ F_{ib}^{2,(t)}$ where \[ F_{ib}^{1,(t)}= \langle \epsilon_i, (\Tilde{\mu}_a-\mu_a^{(t)})-(\Tilde{\mu}_b-\mu_b^{(t)})\rangle\] is the error arising from the GMM part of the model and \[ F_{ib}^{2,(t)} =  \lambda^{(t)}\langle E_{i:}W^{(t)}, (\Tilde{\Pi}_{a:}-\Pi_{a:}^{(t)})-(\Tilde{\Pi}_{b:}-\Pi_{b:}^{(t)})\rangle +\lambda^{(t)}\langle E_{i:}(W-W^{(t)}),\Tilde{\Pi}_{a:}-\Tilde{\Pi}_{b:} \rangle + (\lambda - \lambda^{(t)})\langle E_{i:}W,\Tilde{\Pi}_{a:}-\Tilde{\Pi}_{b:}  \rangle\] is the error coming from the SBM part of the model. We have 
\begin{align*}
F &\leq 2 \max_{\lbrace z^{(t)}: l(z,z^{(t)})\leq \tau^{(0)} \rbrace}\underbrace{\sum_{i=1}^n\max_{b\in [K]\backslash z_i} \frac{(F_{ib}^{1,(t)})^2}{\Delta(z_i,b)^2l(z,z^{(t)})} }_{F_1^{(t)}}+ 2\max_{\lbrace z^{(t)}: l(z,z^{(t)})\leq \tau^{(0)} \rbrace}\underbrace{\sum_{i=1}^n\max_{b\in [K]\backslash z_i} \frac{(F_{ib}^{2,(t)})^2}{ \Delta(z_i,b)^2l(z,z^{(t)})} }_{F_2^{(t)}}
\end{align*} 
and it is sufficient to individually control each term.

\paragraph{Control of $F_1$.} We follow the same steps as in \cite{Gao2019IterativeAF}, the only difference is that we use a different definition for $\Delta$. To begin with,
\begin{align*}
    F_1^{(t)} & \leq \sum_{i=1}^n\sum_{b\in [K]\setminus z_i} \frac{\langle \epsilon_i,(\Tilde{\mu}_{z_i}-\mu_{z_i}^{(t)})-(\Tilde{\mu}_b-\mu_b^{(t)}) \rangle^2}{\Delta(z_i,b)^2l(z,z^{(t)})}\\
        &\leq \sum_{b\in [K]} \sum_{a\in [K]\backslash b} \sum_{i\in \calC_a}  \frac{\langle \epsilon_i,(\Tilde{\mu}_a-\mu_a^{(t)})-(\Tilde{\mu}_b-\mu_b^{(t)}) \rangle^2}{\Delta(a,b)^2l(z,z^{(t)})}\\
        &\leq \sum_{b\in [K]} \sum_{a\in [K]\backslash b} ||\sum_{i\in \calC_a}\epsilon_i\epsilon_i^\top ||\frac{|| (\Tilde{\mu}_a-\mu_a^{(t)})-(\Tilde{\mu}_b-\mu_b^{(t)})||^2}{\Delta(a,b)^2l(z,z^{(t)})}.
\end{align*}
We first need to control $||\sum_{i\in \calC_a}\epsilon_i\epsilon_i^\top ||$ which can be done using the following lemma.
\begin{lemma}
Let $\epsilon_i \overset{i.i.d}{\sim} \calN(0,I_d)$. With probability at least $1-\exp(-0.5n)$, we have \[ ||\sum_{i\in [n]}\epsilon_i\epsilon_i^\top ||\lesssim n+d .\] 
\end{lemma}
\begin{proof}
See Lemma A.2 in \cite{Lu2016StatisticalAC}.
\end{proof}
Next, we need to control $|| \Tilde{\mu}_a-\mu_a^{(t)}||^2$ for all $a\in [K]$, this can be done with the following lemma.
\begin{lemma}
Under the assumptions of Theorem \ref{thm:IRcv}, the following holds with probability at least $1-n^{-\Omega(1)}$. If $z^{(t)}$ satisfies $l(z^{(t)},z)\leq \tau^{(0)} = \frac{\epsilon n\Delta_{min}^2}{K}$ then it implies
\begin{enumerate}
    \item $\max_{k\in [K]} || \tilde{\mu}_k-\mu_k||\lesssim \sqrt{\frac{K(d+\log n)}{n}},$
    \item $\max_{k\in [K]} || \expec(X)^\top( W^{(t)}_{:k}-W_{:k})|| \lesssim \frac{K}{n \Delta_{min}} l(z^{(t)},z)$,
    \item $\max_{k\in [K]} ||(X-\expec(X))^\top W^{(t)}_{:k}|| \lesssim \frac{K\sqrt{(d+n)l(z^{(t)},z)}}{n\Delta_{min}}+\frac{K\sqrt{K(d+\log n)}l(z^{(t)},z)}{n\sqrt{n}\Delta_{min}^2}$
    \item $||\tilde{\mu}_k-\mu_k^{(t)}|| \leq C_3 \frac{K\sqrt{(d+n)l(z^{(t)},z)}}{n\Delta_{min}}$.
\end{enumerate}
%
\end{lemma}

\begin{proof}
Straightforward adaptation the proof of Lemma 4.1 in \cite{Gao2019IterativeAF}.
\end{proof}

By combining the different bounds, we can now conclude that with high probability,  \[ \max_{ \lbrace z^{(t)}: l(z^{(t)},z)\leq \tau^{(0)} \rbrace} F_1 ^{(t)}\lesssim \frac{K^2(Kd/n+1)}{\Delta_{min}^2}\left(1+\frac{K(d/n+1)}{\Delta_{min}^2}\right). \] This quantity goes to zero when $\Delta_{min}^2/K^3 \to +\infty$.

\paragraph{Control of $F_2^{(t)}$.}
Here we can not directly apply the framework developed by \cite{Gao2019IterativeAF}. Different changes are necessary and we need to deal with additional dependencies. 

Let $b \in [K] \neq z_j$, we then have the bound 
\begin{align*}
    (F_{ib}^{2, (t)})^2 &\leq 3  (\lambda^{(t)}\langle E_{i:}W^{(t)}, (\Tilde{\Pi}_{a:}-\Pi_{a:}^{(t)})-(\Tilde{\Pi}_{b:}-\Pi_{b:}^{(t)})\rangle)^2 \\
    &+3(\lambda^{(t)}\langle E_{i:}(W-W^{(t)}),\Tilde{\Pi}_{a:}-\Tilde{\Pi}_{b:} \rangle)^2\\ &
    + 3(\lambda - \lambda^{(t)})^2\langle E_{i:}W,\Tilde{\Pi}_{a:}-\Tilde{\Pi}_{b:}  \rangle^2\\
    &=F_{21}^2+F_{22}^2+F_{33}^2.
\end{align*}
We drop the superscript $(t)$ in the notation for the terms $F_{21}, F_{22}$ and $F_{23}$ for convenience, but clearly they depend on $t$ as well. We will now bound each of the terms $F_{2i}$ for $i=1\ldots 3$ separately. Starting with $F_{21}$, first note that  
\begin{align*}
      |\langle E_{i:}W^{(t)}, (\Tilde{\Pi}_{a:}-\Pi_{a:}^{(t)})-(\Tilde{\Pi}_{b:}-\Pi_{b:}^{(t)})\rangle|^2 &\leq 4||E_{i:}W^{(t)} ||^2\max_{k}||\Tilde{\Pi}_{k:}-\Pi_{k:}^{(t)}||^2.
\end{align*}

With high probability, for all $z^{(t)}$ such that $\lt \leq \tau^{(0)}$ we have
\begin{align*}
     \max_{k\in [K]}||\Tilde{\Pi}_{k:}-\Pi_{k:}^{(t)}||^2 &\leq 2 \max_{k\in [K]}||(W_{:k}-W_{:k}^{(t)})^\top AW||^2+2\max_{k\in [K]}||W_{:k}^{(t) \top}A(W-W^{(t)}) ||^2\\
     & \lesssim \left( \frac{K^{2}\sqrt{p_{max}}\lt}{n^{1.5}\Delta_{min}}\right)^2 \tag{by Lemma \ref{lem:wt}}\\
     &\lesssim \frac{K^4p_{max}\lt^2}{n^3\Delta^2_{min}}
 \end{align*}
  
  and since $\lambda^{(t)}\lesssim \lambda$ (by Lemma \ref{lem:lambdat_ssbm}) it follows that
  \begin{align*}
      \sum_{i=1}^n\max_{b\in [K]\backslash z_i} \frac{F_{21}^2}{\Delta^2(z_i,b)l(z,z^{(t)})} &\lesssim \lambda^2\sum_i ||E_{i:}W^{(t)}||^2\frac{K^4p_{max}\lt^2}{n^3\Delta^2_{min}}\\
      &\lesssim \lambda^2 ||EW^{(t)} ||_F^2\frac{K^4p_{max}\lt}{n^3\Delta^4_{min}}\\
      &\lesssim \lambda^2 K ||EW^{(t)} ||^2\frac{K^4p_{max}\lt}{n^3\Delta^4_{min}}\\
      &\lesssim \lambda^2 K^2 p_{max} \frac{K^4p_{max}\lt}{n^3\Delta^4_{min}} \tag{by a consequence  of Lemma \ref{lem:wt}, fifth item }\\
      &\lesssim \frac{K^4\lt}{n\Delta^4_{min}} \tag{\text{ since } $\lambda \lesssim \frac{n}{Kp_{max}}$}\\
      &\lesssim \frac{K^{3}}{\Delta^2_{min}} \to 0
  \end{align*}
where we used the fact $\frac{K \lt}{n\Delta_{min}^2}\leq \epsilon $ for the last line. Indeed, $\frac{K \lt}{n\Delta_{min}^2} \leq \frac{K \tau^{(0)}}{n\Delta_{min}^2} = \epsilon$.

Let us define $\Delta_2^2(a,b) := \norm{\Pi_{a:}-\Pi_{b:}}^2$ for $a, b \in [K]$.
Since \[ F_{22}^2 \lesssim \lambda^2 ||E_{i:}(W-W^{(t)})||^2||\Tilde{\Pi}_{z_i:}-\Tilde{\Pi}_{b:}||^2 \lesssim \lambda^2 ||E_{i:}(W-W^{(t)})||^2||\Pi_{z_i:}-\Pi_{b:}||^2 \] hence we have w.h.p.  for all $z^{(t)}$ such that $\lt \leq \tau^{(0)}$
\begin{align*}
       \sum_{i=1}^n\max_{b\in [K]\backslash z_i} \frac{F_{22}^2}{\Delta(z_i,b)^2l(z,z^{(t)})} &\lesssim \lambda \sum_i ||E_{i:}(W-W^{(t)})||^2\frac{1}{\lt} \max_{b\in [K]\backslash z_i} \frac{\lambda ||\Pi_{z_i:}-\Pi_{b:}||^2}{\Delta(z_i,b)^2}\\
      &\lesssim ||E(W-W^{(t)})||_F^2\frac{\lambda}{\lt} \tag{because $\Delta^2(z_i,b)\geq \lambda \Delta_2^2(z_i,b)$}\\ 
      &\lesssim K ||E(W-W^{(t)})||^2\frac{\lambda}{\lt}\\
      &\lesssim  \lambda Knp_{max}\frac{K^3\lt}{n^3\Delta_{min}^4} \tag{ by Lemma \ref{lem:wt}}\\
      &\lesssim \frac{K^3}{\Delta_{min}^2} \to 0.
  \end{align*}
Using the same proof technique as in Lemma \ref{lem:ideal_error} \[ (\langle E_{i:}W,\Tilde{\Pi}_{a:}-\Tilde{\Pi}_{b:}  \rangle)^2 \lesssim Kp_{max}\Delta_2^2(a,b)\] holds w.h.p.
Since by Lemma \ref{lem:lambdat_ssbm} we have w.h.p. that for all $z^{(t)}$ such that $\lt \leq \tau^{(0)}$
\[ |\lambda^{(t)}-\lambda|\lesssim \lambda  \frac{K^{2}\lt}{\sqrt{np_{max}}n\Delta_{min}},  \] 
we obtain 
\begin{align*}
     \sum_{i=1}^n\max_{b\in [K]\backslash z_i} \frac{F_{23}^2}{\Delta(z_i,b)^2l(z,z^{(t)})} &\lesssim  \lambda^2 \left(\frac{K^{2}\lt}{\sqrt{np_{max}}n\Delta_{min}}\right)^2\sum_i ||E_{i:}W||^2\max_{b\in [K]\backslash z_i}||\Tilde{\Pi}_{z_i:}-\Tilde{\Pi}_{b:}||^2 \frac{1}{\Delta^2(z_i,b) \lt}\\
     &\lesssim \lambda \frac{K^{4}\lt}{n^3p_{max}\Delta_{min}^2} ||EW||_F^2 \max_{b\in [K]\backslash z_i}\frac{\lambda ||\Pi_{z_i:}-\Pi_{b:}||^2}{\Delta^2(z_i,b)}\\
    &\lesssim \frac{n}{Kp_{max}}\frac{K^{4}\lt}{n^3p_{max}\Delta_{min}^2} Kp_{max}\tag{by Lemma \ref{lem:e}} \\
    &\lesssim \frac{K^4\lt }{n^2p_{max}\Delta^2_{min}} \\
    &\lesssim \frac{K^3}{np_{max}}\\
    &\lesssim  \frac{K^3}{\Delta_{min}^2} \to 0.
\end{align*} 

Consequently, we have established that Condition \ref{cond:f} holds for all $\delta = o(1)$ such that $\delta ^2 = \omega(K^3/\Delta_{min}^2)$.
\end{proof}

\subsection{Error term $G_{ib}^{(t)}$}\label{app:g_error}
In this section we are going to show that the $G$ - error term satisfied condition \ref{cond:h}.
\begin{lemma}\label{lem:g}
 Under the assumptions of Theorem \ref{thm:sirls} (that are also satisfied by Theorem \ref{thm:IRcv}) we have w.h.p. for all $z^{(t)}$ such that $\lt \leq \tau^{(t)}$ \[ \max_{i\in [n]}\max_{b \in [K]\setminus z_i} \frac{|G_{ib}^{(t)}|}{\Delta (z_i,b)^2 } \leq \frac{\delta^{(t+1)}}{8}. \]
\end{lemma}

\begin{proof}

As for $F_{ib}^{(t)}$ we can split $G_{ib}^{(t)} = G_{ib}^{1,(t)} + G_{ib}^{2,(t)}$ where 
\begin{align*}
    G_{ib}^{1,(t)} &= ||\mu_a-\mu_a^{(t)} ||^2-||\mu_a-\tilde{\mu}_a ||^2)-(||\mu_a-\mu_b^{(t)} ||^2-||\mu_a-\tilde{\mu}_b ||^2) \\
    G_{ib}^{2,(t)} &= \lambda^{(t)} (|| P_{i:}W^{(t)}-\Pi_{a:}^{(t)} ||^2-||P_{i:}W^{(t)}-W_{a:}AW^{(t)} ||^2)
    - \lambda^{(t)} (|| P_{i:}W^{(t)}-\Pi_{b:}^{(t)} ||^2-||P_{i:}W^{(t)}-W_{b:}AW^{(t)} ||^2).
\end{align*}
By the proof of Lemma 4.1 in \cite{Gao2019IterativeAF} (last inequality of page 46, equations (115) and (118)), we have w.h.p. for all $z^{(t)}$ such that $\lt \leq \tau^{(t)}$, 
\begin{align*}
    \frac{|G_{ib}^{1,(t)}|}{\Delta^2(a,b)} &\lesssim \left(\frac{K\lt}{n\Delta_{min}}+K\sqrt{\frac{K\lt}{n\Delta_{min}}} \right)^2\Delta_{min}^{-2} \\
    &+\left(\frac{K\lt}{n\Delta_{min}}+K\sqrt{\frac{K\lt}{n\Delta_{min}}} \right)\frac{K\lt}{n\Delta_{min}}\Delta_{min}^{-2} + \left(\frac{K\lt}{n\Delta_{min}}+K\sqrt{\frac{K\lt}{n\Delta_{min}}} \right)\Delta_{min}^{-1}\\
    &\lesssim \frac{K\lt^2}{n^2\Delta_{min}^4}+K\frac{\lt}{n\Delta_{min}^2} \\
    &\lesssim K\frac{\lt}{n\Delta_{min}^2} \tag{since $K\frac{\lt}{n\Delta_{min}^2} < 1$ and $\frac{K\lt^2}{n^2\Delta_{min}^4} < \frac{1}{K}$}  \\
&\lesssim \frac{K \tau^{(t)}}{n \Delta^2_{\min}} \\
&\lesssim \max\left(\frac{7}{8} \frac{K \tau^{(t-1)}}{n \Delta^2_{\min}}, \frac{K \tau}{n \Delta^2_{\min}} \right) \tag{by definition of $\tau^{(t)}$} \\
& \lesssim \max\left(\frac{7}{8} (\frac{8}{7}\epsilon) \delta^{(t)}, \epsilon \delta   \right) \tag{using $\tau = \tau^{(0)} \delta$ and also the definition of $\delta^{(t)}$} \\
&\lesssim \epsilon \delta^{(t)} \tag{since $\delta \leq \delta^{(t)}$}.
\end{align*}
Now by choosing $\epsilon$ to be a suitably small constant ($< 1$), since $\delta^{(t+1)}\leq \frac{7}{8}\delta^{(t)} $ we obtain
\begin{equation*}
    \frac{|G_{ib}^{1,(t)}|}{\Delta^2(a,b)} \leq \frac{\delta^{(t+1)}}{16}.
\end{equation*}

To bound $G_{ib}^{2,(t)}$ we will adapt the method developed in \cite{Han2020ExactCI}.
We have by direct calculation
\begin{align*}
\frac{G_{ib}^{2,(t)}}{\lambda^{(t)}}&= (|| \Pi_{a:}^{(t)}-W_{a:}^\top AW^{(t)}||^2-|| \Pi_{b:}^{(t)}-W_{b:}AW^{(t)}||^2) 
+2\langle P_{i:}W^{(t)} -W_{a:}AW^{(t)}, W_{:a}^\top AW^{(t)}-\Pi_{a:}^{(t)} \rangle \\
&-2\langle P_{i:}W^{(t)} -W_{b:}AW^{(t)}, W_{:b}^\top AW^{(t)}-\Pi_{b:}^{(t)} \rangle \\
& \leq  \left| || \Pi_{a:}^{(t)}-W_{:a}^\top AW^{(t)}||^2-|| \Pi_{b:}^{(t)}-W_{:b}^\top AW^{(t)}||^2 \right| 
+ 4 \max_{a\in [K]} \left|\langle W^\top_{:a}EW^{(t)},(W_{:b}-W_{:b}^{(t)})^\top A W^{(t)} \rangle \right|\\
&+ 2 \left| \langle (\Pi_{a:}-\Pi_{b:})Z^\top W^{(t)},(W_{:b}-W_{:b}^{(t)})^\top A W^{(t)} \rangle \right| \\
&= G_{21}+G_{22}+G_{23}.
\end{align*}
We drop the superscript $(t)$ in the notation for the terms $G_{21}, G_{22}$ and $G_{23}$ for convenience, but clearly they depend on $t$ as well. First observe that w.h.p, it holds for all $z^{(t)}$ such that $\lt \leq \tau^{(t)}$ that
\begin{align*}
    G_{21} \leq \max_{a\in [K]}|| \Pi_{a:}^{(t)}-W_{:a}^\top AW^{(t)}||^2
     = \max_{a\in [K]}|| (W_{a:}^{(t)}-W_{:a})^\top AW^{(t)}||^2
     \lesssim \frac{K^3p_{max}\lt^2}{n^3\Delta_{min}^2}
\end{align*}
where we used Lemma \ref{lem:wt} for the last inequality. This implies by Lemma \ref{lem:lambdat} that 
\begin{align*}
    \max_{b\in [K]\backslash z_i} \frac{\lambda ^{(t)} G_{21}}{\Delta^2 (z_i,b)} \lesssim \frac{K^2 \lt^2}{n^2\Delta^4_{min}} \leq \frac{K^2 (\tau^{(0)}\delta^{(t)})^2}{n^2\Delta^4_{min}}
\end{align*}
%
which implies $ \max_{b\in [K]\backslash z_i} \frac{\lambda ^{(t)} G_{21}}{\Delta^2(z_i,b)} \leq  (\frac{\delta^{(t+1)}}{16})^2$ using the same argument used earlier (for suitably small constant $\epsilon < 1$). 
Next, in order to bound $G_{22}$, note that w.h.p, it holds for all $z^{(t)}$ such that $\lt \leq \tau^{(t)}$ that
\begin{align*}
    G_{22} & \leq 4 \max_{a\in [K]} || W^\top_{:a}EW^{(t)}||\max_{a\in [K]} ||(W_{:b}-W_{:b}^{(t)})^\top A W^{(t)} ||\\
        &\lesssim \frac{K\sqrt{p_{max}}}{\sqrt{n}}\frac{K^{1.5}\sqrt{p_{max}}\lt }{n^{1.5} \Delta_{min}} \tag{by Lemma \ref{lem:wt}}\\
        &\lesssim K^{2.5}\frac{p_{max}}{n} \frac{\lt}{n\Delta_{min}}
\end{align*}
which in turn implies 
\begin{align*}
     \max_{b\in [K]\backslash z_i} \frac{\lambda ^{(t)} G_{22}}{\Delta (z_i,b)^2}& \lesssim \frac{\sqrt{K}}{\Delta_{min}}\frac{K \lt}{n\Delta_{min}^2} \tag{since $\lambda^{(t)} \lesssim n/(Kp_{max})$}\\
     & = o(\delta)
\end{align*}
as $\sqrt{K}/\Delta_{min}\to 0$ and $K\frac{\lt}{n\Delta_{min}^2} < 1$.
    
Finally, in order to bound $G_{23}$, note that w.h.p, it holds for all $z^{(t)}$ such that $\lt \leq \tau^{(t)}$, 
\begin{align*}
    G_{23}&\lesssim  ||\Pi_{a:}-\Pi_{b:})Z^\top W^{(t)}||\max_{b\in [K]} ||(W_{:b}-W_{:b}^{(t)})^\top A W^{(t)}||\\
        &\lesssim \Delta_2(a,b) \frac{K^{1.5}\sqrt{p_{max}}\lt}{n^{1.5}\Delta_{min}}
\end{align*}
which implies
\begin{align*}
    \max_{b\in [K]\backslash z_i} \frac{\lambda ^{(t)} G_{23}}{\Delta (z_i,b)^2}& \lesssim \lambda \frac{\Delta_2(z_i,b)}{\Delta(z_i,b)}\frac{K^{1.5}\sqrt{p_{max}}\lt}{ n^{1.5}\Delta_{min}^2} \\
    &\lesssim \sqrt{\lambda}\frac{\sqrt{Kp_{max}}}{\sqrt{n}} \frac{K\lt}{n\Delta_{min}^2}\\
    &\lesssim \frac{K\lt}{n\Delta_{min}^2} \tag{since $\lambda \lesssim \frac{n}{K p_{max}}$}. 
\end{align*}
For a suitably constant $\epsilon < 1$, this then implies 
\begin{equation*}
    \max_{b\in [K]\backslash z_i} \frac{\lambda ^{(t)} G_{23}}{\Delta (z_i,b)^2} \leq \frac{\delta^{(t+1)}}{16}.
\end{equation*}
\end{proof}

\subsection{Error term $H_{ib}^{(t)}$}\label{app:h_error}
In this section we are going to show that the $H$ - error term satisfied condition \ref{cond:h}.
\begin{lemma}\label{lem:h}
 Under the assumptions of Theorem \ref{thm:sirls} (that are also satisfied by Theorem \ref{thm:IRcv}) we have w.h.p. that for all $z^{(t)}$ such that $\lt \leq \tau^{(t)}$, \[ \max_{i\in [n]}\max_{b \in [K]\setminus z_i} \frac{|H_{ib}^{(t)}|}{\Delta (z_i,b)^2 } \leq \frac{\delta^{(t+1)}}{8}. \]
\end{lemma}

\begin{proof}

As before, we can split $H_{ib}^{(t)} = H_{ib}^{1,(t)} + H_{ib}^{2,(t)}$ where
%
\begin{align*}
     2H_{ib}^{1,(t)} & =||\mu_a-\tilde{\mu}_a||^2-||\mu_a-\tilde{\mu}_b ||^2+ ||\mu_a-\mu_b||^2 \\
     2H_{ib}^{2,(t)} &=\lambda^{(t)} (||P_{i:}W^{(t)}-W_{a:}^\top AW^{(t)} ||^2 - ||P_{i:}W^{(t)}-W_{b:}^\top AW^{(t)} ||^2 + ||\Pi_{a:}-\Pi_{b:}||^2)+ (\lambda-\lambda^{(t)})||\Pi_{a:}-\Pi_{b:}||^2.
\end{align*}

By an immediate adaptation of Lemma 4.1 in \cite{Gao2019IterativeAF} it holds w.h.p. that for all $z^{(t)}$ such that $\lt \leq \tau^{(t)}$,
\[ \frac{|H_{ib}^{1,(t)}|}{\Delta (z_i,b)^2}\lesssim \frac{K(d+\log n)}{n\Delta_{min}^2}+\sqrt{\frac{K(d+\log n)}{n\Delta_{min}^2}}\to 0 \] 
as long as $K/\Delta_{min}^2\to 0$.

It remains to uniformly control  $H_{ib}^{2,(t)}$, let us split it\footnote{We drop the superscript $(t)$ in the notation for $H_1, H_2$ for convenience, but clearly they both depend on $t$ as well.} as $H_{ib}^{2,(t)} = \lambda^{(t)} H_1 + H_2$. First note that by Lemma \ref{lem:lambdat_ssbm}
it holds w.h.p. that for all $z^{(t)}$ such that $\lt \leq \tau^{(t)}$,
\begin{align*}
     H_2: =(\lambda-\lambda^{(t)})||\Pi_{a:}-\Pi_{b:}||^2
     \lesssim \lambda \frac{K\lt}{n\Delta_{min}^2}  \Delta^2_2(a,b)
\end{align*}
which implies
\begin{align*}
    \frac{|H_2|}{\Delta (z_i,b)^2} \lesssim \frac{K\lt}{n\Delta^2_{min}}.
\end{align*}  
Then, we obtain $ \frac{|H_2|}{\Delta (z_i,b)^2} \leq \frac{\delta^{(t+1)}}{16}$ for a small enough constant $\epsilon < 1$.

Now observe that 
\begin{align*}
    H_1:&=||P_{i:}W^{(t)}-W_{a:}^\top AW^{(t)} ||^2 - ||P_{i:}W^{(t)}-W_{b:}^\top AW^{(t)} ||^2 + ||\Pi_{a:}-\Pi_{b:}||^2 \\
    &= ||W_{a:}^\top EW^{(t)} ||^2  + (||\Pi_{a:}-\Pi_{b:}||^2 - ||P_{i:}W^{(t)}-W_{b:}^\top PW^{(t)} ||^2 )\\
    &-(||P_{i:}W^{(t)}-W_{b:}^\top AW^{(t)} ||^2-||P_{i:}W^{(t)}-W_{b:}^\top PW^{(t)} ||^2)\\
    &= (||\Pi_{a:}-\Pi_{b:}||^2 - ||P_{i:}W^{(t)}-W_{b:}^\top PW^{(t)} ||^2 )+ (||W_{a:}^\top EW^{(t)} ||^2-||W_{b:}^\top EW^{(t)} ||^2)\\
    &+2\langle P_{i:}W^{(t)}-W_{b:}^\top PW^{(t)} ,W_{:b}^\top E_{}W^{(t)} \rangle \\
    &= H_{11}^{(t)}+H_{12}^{(t)}+H_{13}^{(t)}.
\end{align*} 

By writing \[P_{i:}W^{(t)}-W_{b:}^\top PW^{(t)} = (\Pi_{a:}-\Pi_{b:})Z^\top W^{(t)}, \] we obtain w.h.p. that for all $z^{(t)}$ such that $\lt \leq \tau^{(t)}$, 
\begin{align*}
    |H_{13}^{(t)}| &\lesssim || P_{i:}W^{(t)}-W_{b:}^\top PW^{(t)}|| ||W_{:b}^\top E_{}W^{(t)}||
    \lesssim ||\Pi_{a:}-\Pi_{b:} ||\frac{K\sqrt{p_{max}}}{\sqrt{n}}
    \lesssim ||\Pi_{a:}-\Pi_{b:} ||^2 \frac{K}{\sqrt{np_{max}}}.
\end{align*}
In particular, 
\[ \frac{\lambda^{(t)}|H_{13}^{(t)}|}{\Delta^2(z_i,b)} \lesssim \frac{\lambda \Delta_2^2(z_i,b) K}{\Delta^2(z_i,b) \sqrt{np_{max}}}  \lesssim  \frac{K}{\sqrt{np_{max}}} \to 0.\]
Next observe that w.h.p., it holds for all $z^{(t)}$ such that $\lt \leq \tau^{(t)}$ that
\begin{align*}
    \abs{||W_{a:}^\top EW^{(t)} ||^2-||W_{b:}^\top EW^{(t)} ||^2} \leq \max_{k\in [K]}||W_{k:}^\top EW^{(t)} ||^2
    {\lesssim } \frac{K^2p_{max}}{n}  
\end{align*}
where the last inequality uses Lemma \ref{lem:wt}. This implies
\[ 
\frac{\lambda^{(t)}|H_{12}^{(t)}|}{\Delta^2(z_i,b)} \lesssim \frac{K}{\Delta^2_{min}} \to 0.
\] 

Finally, it remains to bound $|H_{11}^{(t)}|$. To begin with,
\[ 
|H_{11}^{(t)}| := \abs{ ||\Pi_{a:}-\Pi_{b:}||^2 - ||P_{i:}W^{(t)}-W_{b:}^\top PW^{(t)} ||^2 } = \abs{||\Pi_{a:}-\Pi_{b:}||^2 - ||(\Pi_{a:}-\Pi_{b:}) Z^\top W^{(t)} ||^2 }.
\]
Using the fact 
\[
\abs{||\Pi_{a:}-\Pi_{b:}||^2 -||(\Pi_{a:}-\Pi_{b:}) Z^\top W^{(t)} ||^2} \leq \left(\norm{Z^\top W^{(t)} - I}^2 + 2\norm{Z^\top W^{(t)} - I}\right) \norm{\Pi_{a:}-\Pi_{b:}}^2,
\]
we obtain by the proof of part $1$ of Lemma \ref{lem:wt} that w.h.p., it holds for all $z^{(t)}$ such that $\lt \leq \tau^{(t)}$
\begin{align*}
    |H_{11}^{(t)}| \lesssim ||\Pi_{a:}-\Pi_{b:}||^2\frac{K\lt }{n\Delta_{min}^2} 
\end{align*}
which implies  
\begin{align*}
    \frac{\lambda^{(t)}|H_{11}^{(t)}|}{\Delta^2(z_i,b)} &\lesssim \frac{K\lt }{n\Delta_{min}^2}.
 \end{align*}
Then as before, for a suitably small constant $\epsilon < 1$, this implies
\begin{align*}
    \frac{\lambda^{(t)}|H_{11}^{(t)}|}{\Delta^2(z_i,b)} \leq \frac{\delta^{(t+1)}}{16}.
\end{align*}

By summing all these inequalities we see that $G^{(t)}_{ib}$ and $H^{(t)}_{ib}$ satisfy  Condition \ref{cond:h}.
\end{proof}

\subsection{Conclusion}\label{app:thm2_cl}
By Lemma \ref{lem:f}, \ref{lem:g} and \ref{lem:h} the Conditions \ref{cond:f} and \ref{cond:h} are satisfied.
In order to apply Theorem \ref{thm:gao_ext}, we also need to show that Condition \ref{cond:ideal} is satisfied. To this end, let us define the events 
\[ 
\Omega(\delta^{(t)} ) := \lbrace C_i(a,b) \leq -\left(\frac{1-\delta^{(t)}}{2}\right)\Delta^2(a,b)\rbrace, \quad t \geq 1, 
\] 
where we recall that \[ C_i(a,b)=\langle \epsilon_i, \Tilde{\mu}_a-\Tilde{\mu}_b\rangle+ \lambda \langle E_{i:}W, \Tilde{\Pi}_{a:}-\Tilde{\Pi}_{b:}\rangle .\]
Note that by definition of $\delta^{(t)}$, we have for all $t \geq 1$ 
\begin{equation} \label{eq:delta_t_1_rel}
    \Omega(\delta^{(t)} ) \subseteq \Omega(\delta^{(1)}) \implies \indic_{\Omega(\delta^{(t)} )} \leq \indic_{\Omega(\delta^{(1)})} \implies \xi(\delta^{(t)}) \leq \xi(\delta^{(1)}).
\end{equation}
Hence it suffices to bound $\xi(\delta^{(1)})$. To achieve this, it is crucial to bound $\prob (\Omega (\delta^{(1)}))$ since the bound on $\xi(\delta^{(1)})$ then follows via an easy adaptation of Lemma \ref{lem:ideal_error}.

We can bound $\prob (\Omega (\delta^{(1)}))$ by an immediate adaptation of Lemma \ref{lem:ideal_error_ssbm} -- we just need to replace $\Delta^2/4$ by $(1-\delta^{(1)})\Delta^2/4$ in the last step of the lemma, leading to
\[ 
 \prob (\Omega (\delta^{(1)}))\leq \exp (-(1-\delta^{(1)} +o(1))\tilde{\Delta}^2). 
\] 
Then as mentioned earlier, an easy adaptation of Lemma \ref{lem:ideal_error} along with \eqref{eq:delta_t_1_rel} implies that w.h.p., it holds for all $t\geq 1$ that
\[ 
\xi(\delta^{(t)}) \leq n\exp (-(1-\delta^{(1)} +o(1))\tilde{\Delta}^2).
\] 
But we know that \[n\exp (-(1-\delta^{(1)} +o(1))\tilde{\Delta}^2) \leq \frac{3}{4}\tau =\frac{3}{4}\tau^{(0)}\delta \] 
for $n$ large enough, by a suitable choice of $\delta = o(1)$, and the fact that $\tilde{\Delta}^2\asymp \log n/K \to \infty$. Hence the assumptions of Theorem \ref{thm:gao_ext} are satisfied. Moreover, for all $t\gtrsim \log (1/\delta)$, Corollary \ref{cor:contraction} yields the bound  \[ \lt \lesssim \xi(\delta)+ \tau^{(0)}(1/8)^{t-\Theta (\log(1/\delta))}\lesssim ne^{-(1+o(1))\tilde{\Delta}^2} + \tau^{(0)}(1/8)^{t-\Theta (\log(1/\delta))}. \] Since $\tau^{(0)} \lesssim n\log n/K$ and $\delta$ can be chosen such that $\log(1/\delta) \lesssim \log n$, we obtain that for $t\gtrsim \log n$ \[ \lt \leq  ne^{-(1+o(1))\tilde{\Delta}^2}.\] 



\section{Proof of Theorem \ref{thm:minimax}}\label{app:minimax}

\begin{proof} The proof follows the same steps as in the proof of Theorem 3.3 in  \cite{Lu2016StatisticalAC}, only the last part needs to be changed. For the sake of completeness, we reproduce the arguments below.
Let us denote  \begin{equation}
    \label{eq:ham}h(z',z'')=\sum_{i\in [n]} \indic_{\lbrace z'(i)\neq z''(i)\rbrace}
\end{equation}  to be the unnormalized Hamming distance between $z',z'' \in [K]^n$.
Without lost of generality we can assume that \[ \min_{k,k'}||\mu_k-\mu_{k'}||=||\mu_1-\mu_{2}||.\] For each $k \in [K]$, let $T_k$ a subset of $\calC_k$ with cardinality $\frac{3n}{4K}$. Define $T=\cup_{k=1}^KT_k$ and \[ \mathcal{Z} = \lbrace \hat{z} : \hat{z}_i=z_i \text{ for all }i\in T \rbrace.\]

For all $\hat{z}\neq \Tilde{z} \in \mathcal{Z}$ we have 
\[ \frac{h(\hat{z},\Tilde{z})}{n} \leq \frac{1}{4}
\] 
and for all permutations $\sigma \in \mathfrak{S}_K, \sigma \neq Id$ ( where $Id$ denotes the identity permutation) we have \[\frac{h(\sigma(\hat{z}),\Tilde{z})}{n}\geq \frac{1}{2} .\] Thus $r(\hat{z},\Tilde{z}) = \frac{h(\hat{z},\Tilde{z})}{n}$. Then following the same arguments as in the proof of Theorem 2 in \cite{DCBM2016} we can obtain \begin{equation}\label{eq:minmax1}
    \inf_{\hat{z}} \sup_{\theta \in \Theta} \expec (r(\hat{z},z)) \geq \frac{1}{6|T^c|}\sum_{i \in T^c}\frac{1}{2K^2} \inf_{\hat{z}_i} (\prob_1(\hat{z}_i=2)+\prob_2(\hat{z}_i=1))
\end{equation}  
where $\prob_k$ denotes the probability distribution of the data when $z_i=k$. By the Neyman Pearson Lemma, the infinimum of the right hand side of \eqref{eq:minmax1} is achieved by the likelihood ratio
test. From  Section 3.1 in \cite{MiniMaxZhang2016},  the log-likelihood of the SBM part can be rewritten as 
\[ 
\log \left(\frac{p(1-q)}{q(1-p)} \right)\sum_{i<j}A_{ij}\indic_{\lbrace z_i=z_j\rbrace}+ f(A) 
\] 
where $f(A)$ doesn't depend on $z$. Consequently,  
\begin{align}\label{eq:minimax2}
        &\frac{1}{2} \inf_{\hat{z}_i}\left(\prob_1(\hat{z}_i=2)+\prob_2(\hat{z}_i=1) \right) \nonumber \\ 
        &= \prob \left(-0.5||\epsilon_i||^2+ \log\left(\frac{p(1-q)}{q(1-p)}\right)\sum_{j\in \calC_1}A_{ij}\leq -0.5||\mu_1 + \epsilon_i -\mu_2 ||^2 
        +\log\left(\frac{p(1-q)}{q(1-p)}\right)\sum_{j\in \calC_2}A_{ij} \right)
\end{align}
Let us denote $Z_i = \log(\frac{p(1-q)}{q(1-p)})(\sum_{j\in \calC_2}A_{ij} - \sum_{j\in \calC_1}A_{ij})$, this is a random variable independent of $\epsilon_i$. So we get
\begin{align*}
   (\ref{eq:minimax2})&= \prob (0.5||\mu_1-\mu_2||^2-Z_i \leq -\langle \epsilon_i, \mu_1-\mu_2 \rangle)\\
    &\geq  \prob (0.5||\mu_1-\mu_2||^2-Z_i \leq -\langle \epsilon_i, \mu_1-\mu_2 \rangle \mid Z_i> 0)\prob(Z_i > 0)\\
    &\geq  \prob (||\mu_1-\mu_2||^2\leq -2\langle \epsilon_i, \mu_1-\mu_2 \rangle)\prob(Z > 0)\\
    & \geq \exp\left(-\frac{\Delta_1^2}{8}\right)\exp\left(-n\frac{(1+o(1))(\sqrt{p}-\sqrt{q})^2}{K}\right) \\
    &\geq \exp(-(1+o(1))\Tilde{\Delta}^2).
\end{align*}
Here we used for the penultimate inequality a result from the proof of Theorem 3.3 in \cite{Lu2016StatisticalAC} and also use Lemma 5.2 in \cite{MiniMaxZhang2016}.
\end{proof}

\section{Technical Lemmas}\label{app:lemmas}
In this section we provide some useful inequalities which are used frequently in the proofs of our main results.
\subsection{General concentration inequalities}
\begin{lemma}[Chernoff simplified bound]\label{lem:chern_bin} Let $X_1, \ldots, X_n \overset{\text{ind.}}{\sim} \mathcal{B}(p)$. Then \[ \prob\left(\abs{\frac{1}{n}\sum_{i \in [n]} X_i-p}\geq t \right) \leq \exp(-2nt^2).\] 
\end{lemma}

\begin{lemma}
\label{lem:e} Assume that $A \sim SBM(Z,\Pi)$. Let $E=A-\expec(A)$.
Then with probability at least $1-n^{-\Omega(1)}$ the following holds.
\begin{enumerate}
    \item $||E||\leq \sqrt{np_{max}}$,
    \item $||EW||_F^2 \lesssim   K^2 p_{max}$.
\end{enumerate}
\end{lemma}
\begin{proof}
The first inequality is a classical result used for SBM in the relatively sparse regime $p_{max}=\omega (\log n)$. It can obtained as a consequence of Remark 3.13 in \cite{bandeira2016}. The second inequality follows from \begin{align*}
    ||EW||_F^2 &\leq  K ||EW||^2\leq K||E||^2||W||^2\lesssim K^2p_{max}.
\end{align*}
\end{proof}

\subsection{Concentration rate for the ideal oracle error under CSSBM}
\begin{lemma}\label{lem:ideal_error_ssbm}
Recall that \[\Omega_1 = \left\lbrace \langle \epsilon_i, \mu_{z_i}-\mu_b\rangle+ \lambda \langle E_{i:}W, \Pi_{z_i:}-\Pi_{b:}\rangle \leq \frac{-(1-\delta-\Bar{\delta})\Delta^2(z_i,b)}{2} \right\rbrace \] and suppose that $\delta, \bar{\delta} = o(1)$. Then under the assumptions of Theorem \ref{thm:IRcv}, we have  \[  \prob(\Omega_1) \leq \exp (-(1+o(1))\Tilde{\Delta}^2 ) \] where \[ \Tilde{\Delta}^2 = \frac{1}{8}\min_{k, k'} ||\mu_k-\mu_{k'} ||^2+\frac{\log n }{K}(\sqrt{p'}-\sqrt{q'})^2.\]
\end{lemma}

\begin{proof}
We are going to bound the m.g.f of $Z =\langle \epsilon_i, \mu_{z_i}-\mu_b\rangle+ \lambda \langle E_{i:}W, \Pi_{z_i:}-\Pi_{b:}\rangle$ and use Chernoff method.  We have for all $t<0$
\begin{align*}
    \log \expec e^{tZ} & \leq \log \expec e^{t\langle \epsilon_i, \mu_{z_i}-\mu_b\rangle} + \log \expec e^{t\lambda \langle E_{i:}W, \Pi_{z_i:}-\Pi_{b:}\rangle} \tag{by independence} \\
        & \leq ||\mu_{z_i}-\mu_b||^2\frac{t^2}{2}+\frac{n}{K}\log (pe^{t\lambda(p-q)K/n }+1-p)(qe^{-t\lambda (p-q)K/n}+1-q)\\
        &-t\lambda (p-q)^2.
\end{align*}

By choosing $t=-1/2$ we get \[ e^{t\lambda(p-q)K/n } = \sqrt{\frac{q(1-p)}{p(1-q)}}\]
and thus \[ \log (pe^{t\lambda(p-q)K/n }+1-p)(qe^{-t\lambda (p-q)K/n}+1-q) = \log (pq +(1-p)(1-q) + 2 \sqrt{pq}\sqrt{(1-p)(1-q)}).\] This last quantity is equal to $-(1+o(1))(\sqrt{p}-\sqrt{q})^2$.
We can now conclude by remarking that \[ \prob(\Omega_1)\leq \prob \left(-\frac{1}{2}\langle \epsilon_i, \mu_{z_i}-\mu_b\rangle- \frac{1}{2}\lambda \langle E_{i:}W, \Pi_{z_i:}-\Pi_{b:}\rangle \geq (1+o(1))\frac{\Delta_{min}^2}{4}\right)\]
hence
\begin{align*}
    \prob (\Omega_1) & \leq \expec e^{-\frac{Z}{2}-\frac{\Delta_{min}^2}{4}}\\
    & \leq \exp(\frac{||\mu_{z_i}-\mu_b ||^2}{8}- \frac{n}{K}(1+o(1))(\sqrt{p}-\sqrt{q})^2+ \frac{\lambda}{2}(p-q)^2 -(1+o(1))\frac{\Delta^2_{min}}{4})\\
    & \leq \exp (-(1+o(1))\Tilde{\Delta}^2 ),
\end{align*}
since $\Delta_{min}^2 = ||\mu_{z_i}-\mu_b ||^2 +2\lambda (p-q)^2 $.
\end{proof}

\subsection{Concentration rate for the ideal oracle error under the general setting}
In the general setting it is more difficult to derive a sharp concentration inequality for the oracle error. Here we use gaussian approximation, but it leads to a slightly sub-optimal convergence rate. 

\begin{lemma}\label{lem:ideal_error_csbm}  Recall the definition of $\Omega_1$ from the previous lemma. Under the assumptions of Theorem \ref{thm:sirls} we have \[  \prob(\Omega_1) \leq \exp \left(-\frac{1}{8}\Delta_{min}^2 \right). \] 
\end{lemma}
\begin{proof}
First observe that \begin{align*}
     t\lambda \langle E_{i:}W, \Pi_{z_i:}-\Pi_{b:}\rangle &=t\lambda \sum_{k\in [K]} (\Pi_{z_ik}-\Pi_{bk})\frac{\sum_{j\in \calC_k}E_{ij}}{n_k}\\
     &= \sum_{k\in [K]} t\lambda (\Pi_{z_ik}-\Pi_{bk})\frac{\sum_{j\in \calC_k}A_{ij}-\Pi_{z_ik}}{n_k}.
\end{align*} 
The sum over $k$ involves independent random variables so in order to bound the $m.g.f.$ of $\langle E_{i:}W, \Pi_{z_i:}-\Pi_{b:}\rangle $ it is sufficient to control the $m.g.f.$ of $\sum_{j\in \calC_k}A_{ij}-\Pi_{z_ik}$ for each $k$. Setting $t' = \lambda t\frac{|\Pi_{z_ik}-\Pi_{bk}|}{n_k}$, we have \begin{align*}
   \log \expec (e^{t'\sum_{j\in \calC_k}(A_{ij}-\Pi_{z_ik})})&= \log (\Pi_{z_ik}e^{t'}+1-\Pi_{z_ik}) - n_kt'\Pi_{z_ik}\\
   &\leq n_k\Pi_{z_ik}(e^{t'}-t'-1)\\
   &\leq n_k\Pi_{z_ik} \frac{e(t')^2}{2} \tag{ by Taylor-Lagrange inequality}\\
   &\leq 1.5n_kp_{max}(\lambda t\frac{\Pi_{z_ik}-\Pi_{bk}}{n_k})^2\\
   &\leq 1.5\lambda |\Pi_{z_ik}-\Pi_{bk}|^2t^2.
\end{align*}
 For the second inequality we used the fact that for $0<x<1$, $\log(1-x)\leq -x$.
 
 Consequently, \[ \log \expec e^{tZ} \leq ||\mu_{z_i}-\mu_b ||^2\frac{t^2}{2}+ 1.5\lambda ||\Pi_{z_i:}-\Pi_{b:} ||^2t^2\] and \begin{align*}
     \prob (\Omega_1) &\leq e^{ ||\mu_{z_i}-\mu_b ||^2\frac{t^2}{2}+ 1.5\lambda ||\Pi_{z_i:}-\Pi_{b:} ||^2t^2- \frac{\Delta^2(z_i,b)}{4}}.
 \end{align*}
 For $t=1/2$ we then obtain \[ \prob (\Omega_1) \leq e^{-\frac{ ||\mu_{z_i}-\mu_b ||^2}{8}-\frac{\lambda}{8}||\Pi_{z_i:}-\Pi_{b:} ||^2} \leq e^{-\frac{1}{8}\Delta_{min}^2} .\]
\end{proof}

\subsection{Useful inequalities to control the error terms}
%
Recall the Hamming distance $h$ defined in \eqref{eq:ham}.
\begin{lemma}\label{lem:ht}
For all $z, z' \in [K]^n$ we have \[ h(z,z') \leq \frac{l(z,z')}{\Delta_{min}^2}.\]
\end{lemma}
\begin{proof}
\[ \sum_{i \in [n]} \indic_{z_i \neq z'_i} \leq \sum_{i \in [n]} \frac{\Delta(z_i, z_i')^2}{\Delta_{min}^2}\indic_{z_i \neq z_i'} =  \frac{l(z,z')}{\Delta_{min}^2}. \]
\end{proof}

\begin{lemma}\label{lem:nt} Assume that for some $\alpha >1$ \[  \frac{n}{\alpha K} \leq n_k \leq \frac{\alpha n}{K}.\] 
If $l(z,z^{(t)}) \leq  n\Delta_{min}^2/(2\alpha K)$ then for all $k \in [K]$ \[  \frac{ n}{2\alpha K} \leq n_k^{(t)} \leq \frac{2\alpha n}{K}.\]
\end{lemma}
\begin{proof}
Since for all $k \in [K]$ we have $n/(\alpha  K) \leq n_k \leq \alpha n/K$, \begin{align*}
    \sum_{i \in \calC_k^{(t)}}1 &\geq \sum_{i \in \calC_k \cap \calC_k^{(t)}}1 \geq \sum_{i \in \calC_k}1-\sum_{i \in [n]}\indic_{z_i \neq z_i^{(t)}}\\
    & \geq  \frac{n}{\alpha K} - h(z,z^{(t)}) \overset{\text{Lemma }\ref{lem:ht}}{\geq } \alpha \frac{n}{K} - \frac{l(z,z^{(t)})}{\Delta_{min}^2}\\
    &\geq  \frac{\alpha n}{2K}
\end{align*}
by assumption. The other inequality is proved in a similar way.
\end{proof}

\begin{lemma}\label{lem:wt}Assume that $A \sim SBM(Z,\Pi)$ with equal size communities and suppose that the conditions of Theorem \ref{thm:sirls} are satisfied\footnote{These assumptions are clearly satisfied by Theorem \ref{thm:IRcv} as well.}. Then with probability at least $1-n^{-\Omega(1)}$ the following holds for all $z^{(t)}$ such that $\lt \leq \tau^{(0)}$.
\begin{enumerate}
    \item $ \max_{k\in [K]}|| W^{(t)}_{:k}-W_{:k}|| \lesssim \frac{K^{1.5}}{n^{1.5}\Delta_{min}^2} l(z^{(t)},z),$
    \item $\max_{k\in [K]}|| (W_{:k}^{(t)}-W_{:k})^\top AW|| \lesssim  \frac{K^{1.5}\sqrt{p_{max}}}{n^{1.5}\Delta_{min}}l(z, z^{(t)})$,
    \item $ \max_{k\in [K]}|| W_{:k}^{(t)\top} A(W-W^{(t)})||\lesssim \frac{K^{2}\sqrt{p_{max}}\lt}{n^{1.5}\Delta_{min}},$
    \item $ \max_{k\in [K]} || (W_{:k}^{(t)}-W_{:k})^\top AW^{(t)}|| \lesssim   \frac{K^{1.5}\sqrt{p_{max}}\lt}{n^{1.5}\Delta_{min}}$,
    \item $||Z^\top W^{(t)}|| \lesssim 1 $.
\end{enumerate}
\end{lemma}

\begin{proof}
This is a rather straightforward adaptation of Lemma 4 in \cite{Han2020ExactCI}, but for completeness we include a proof adapted to our setting with our notations.

\paragraph{Proof of 1.} First observe that $Z$ is rank $K$ and $\lambda_K(Z)= \sqrt{n_{min}}$ so that \[ || W^{(t)}_{:k}-W_{:k}|| \leq n_{min}^{-1/2} ||I-Z^\top W^{(t)} ||.\] For any $k \in [K]$, denote $\delta_k=1-(Z^\top W^{(t)})_{kk}$. Since for all $k, k' \in [K]$ \[ (Z^\top W^{(t)})_{kk'}=\frac{\sum_{i \in \calC_k}\indic_{z_i^{(t)}=k'}}{n_{k'}^{(t)}},\] we have \[ 0 \leq \delta_k \leq 1, \quad \sum_{k'\in [K]\backslash k} (Z^\top W^{(t)})_{k'k} = \delta_k .\] Therefore, 
\begin{align*}
    ||Z^\top W^{(t)}-I || & = \sqrt{\sum_{k \in [K]} \left( \delta_k^2+ \sum_{k'\in [K]\backslash k} (Z^\top W^{(t)})_{k'k}^2 \right)}\\
    & \leq \sqrt{\sum_{k \in [K]} \left( \delta_k^2+ \left(\sum_{k'\in [K]\backslash k} (Z^\top W^{(t)})_{k'k}\right)^2 \right)}\\
    & \leq \sqrt{2\sum_{k \in [K]} \delta_k^2} \leq \sqrt{2}\sum_{k \in [K]}\delta_k \\
    & = \sqrt{2} \sum_{k\in [K]}\frac{\sum_{i \in \calC^{(t)}_k} \indic_{z_i \neq k}}{n_k^{(t)}}\\
    & \leq \sqrt{2}\max_k(n_k^{(t)})^{-1} \sum_{i \in [n]}\indic_{z_i \neq z_i^{(t)}}\\
    & \overset{\text{Lemma \ref{lem:nt}}}{\lesssim } \frac{K}{n}h(z,z^{(t)}) \overset{\text{Lemma \ref{lem:ht}} }{\lesssim }  K\frac{l(z,z^{(t)})}{n \Delta_{min}^2} \numberthis \label{eq:izw}.
\end{align*}

\paragraph{Proof of 2.} Observe that with probability at least $1-n^{-\Omega(1)}$ we have 
\begin{align*}
    \max_{k\in [K]}|| (W_{:k}^{(t)}-W_{:k})^\top AW|| &\leq  \max_{k\in [K]}|| (W_{:k}^{(t)}-W_{:k})^\top PW|| +   \max_{k\in [K]}|| (W_{:k}^{(t)}-W_{:k})^\top EW|| \\ 
    &\leq  \max_{k\in [K]}|| (W_{:k}^{(t)}-W_{:k})^\top Z\Pi|| +  || EW|| \max_{k\in [K]}|| (W_{:k}^{(t)}-W_{:k})|| \\
    &\leq ||\Pi_{b:}-\sum_{j\in \calC_b^{(t)}}\frac{\Pi_{z_j:}}{n^{(t)}_b} ||+ C \sqrt{Kp_{max}}\max_{k\in [K]}||(W_{:k}^{(t)}-W_{:k})||\\
    &\lesssim  ||\Pi_{b:}-\sum_{j\in \calC_b^{(t)}}\frac{\Pi_{z_j:}}{n^{(t)}_b} ||+ \frac{K^2\sqrt{p_{max}}}{n^{1.5}\Delta_{min}^2}\lt. 
\end{align*}  
Recall that $\Delta_2^2(a,b) := \norm{\Pi_{a:}-\Pi_{b:}}^2$.
Then we have 
 \begin{align*}
    \norm{\Pi_{b:}-\sum_{j\in \calC_b^{(t)}}\frac{\Pi_{z_j:}}{n^{(t)}_b}} & = \norm{ \sum_{\substack{j\in \calC_b^{(t)}\\ b'\in [K]\backslash b}}\frac{\indic_{\lbrace z_j = b'\rbrace }}{n_b^{(t)}}(\Pi_{b:}-\Pi_{b':})} \\
    &\leq C\frac{K}{ n}\sum_{\substack{j\in \calC_b^{(t)}\\ b'\in [K]\backslash b}}\max_{b,b'}\Delta_2(b,b')\indic_{\lbrace z_j = b'\rbrace }\\
    &\leq  C\frac{K}{ n}\max_{b,b'}\Delta_2(b,b') h(t,t^{(t)})\\
    &\leq C\frac{K \Delta_{min}}{\sqrt{\lambda} n \Delta^2_{min}}l(z, z^{(t)}) \tag{ since $\max_{b,b'} \Delta_2(b,b')\lesssim \frac{\Delta_{min}}{\sqrt{\lambda}}$ for SBM }\\
    & \leq C \frac{K^{1.5}\sqrt{p_{max}}}{n^{1.5}\Delta_{min}}l(z, z^{(t)}).
\end{align*}
    Consequently, by summing the previous bounds and using the first inequality of the Lemma we get \[  \max_{k\in [K]}|| (W_{:k}^{(t)}-W_{:k})^\top AW|| \lesssim \frac{K^{1.5}\sqrt{p_{max}}}{n^{1.5}\Delta_{min}}l(z, z^{(t)})+ \frac{K^2\sqrt{p_{max}}}{n^{1.5}\Delta_{min}^2}\lt .\] In our setting $\Delta^2_{min} \asymp \log n$ so the first term is dominant.

    \paragraph{Proof of 3.} First let's bound $ \max_{k\in [K]}|| W_{:k}^{(t)\top} P(W-W^{(t)})||$. By Lemma \ref{lem:nt} we have $||W_k^{(t)} || \lesssim  \sqrt{K/n}$, so \begin{align*}
        \max_{k\in [K]}|| W_{:k}^{(t)\top} P(W-W^{(t)})|| &\leq  \max_{k\in [K]} || W_{:k}^{(t)\top} Z|| ||\Pi Z^\top (W-W^{(t)})||\\
        & \lesssim  ||\Pi Z^\top (W-W^{(t)}) ||_F\\
        & \lesssim \sqrt{K} \max_{k\in [K]}||(W_{:k}^{(t)}-W_{:k})^\top Z \Pi ||\\
        & \lesssim \frac{K^{2}\sqrt{p_{max}}}{n^{1.5}\Delta_{min}}l(z, z^{(t)}). \tag{by the proof of part 2}
    \end{align*} 
    
     We now give an upper bound for $ \max_{k\in [K]}|| W_{:k}^{(t)\top} E(W-W^{(t)})||$. By triangle inequality, \[ || W_{:k}^{(t)\top} E(W-W^{(t)})|| \leq || W_{:k}^{\top} E(W-W^{(t)})|| +|| (W_{:k}^{(t)}-W_{k:})^\top E(W-W^{(t)})||.\]
     First we have \begin{align*}
         || W_{:k}^{\top} E(W-W^{(t)})|| &\leq  || W_{:k}|| || E|| ||(W-W^{(t)})||\\
            &\lesssim \frac{K^2\sqrt{p_{max}}}{n^{1.5}\Delta_{min}^2} l(z^{(t)},z).
     \end{align*}
    On the other hand, we also have \begin{align*}
        || (W_{:k}^{(t)}-W_{k:})^\top E(W-W^{(t)})|| &\leq ||W_{:k}-W_{:k}^{(t)} |||| E(W-W^{(t)})||\\
            &\leq ||E||\sqrt{K}\max_{k}||W_{:k}-W_{:k}^{(t)} || ^2 \\
            & \lesssim \frac{K^{3.5}\sqrt{np_{max}}\lt}{n^2\Delta^2_{min}}\frac{\lt}{n\Delta_{min}^2}\\
            &\lesssim \frac{K^{3.5}\sqrt{p_{max}}\lt}{n^{1.5}\Delta^2_{min}}
    \end{align*}
where the last inequality comes from the fact that by assumption $l(z,z^{(t)})\leq \tau \leq \epsilon \frac{n\Delta_{min}^2}{K}$.

Thus it follows that
\[  \max_{k\in [K]}|| W_{:k}^{(t)\top} P(W-W^{(t)})|| \lesssim \frac{K^{2}\sqrt{p_{max}}\lt}{n^{1.5}\Delta_{min}}.
\]
    
    \paragraph{Proof of 4.} First note that \begin{align*}
         || (W_{:k}^{(t)}-W_{:k})^\top PW^{(t)}||&\leq ||(W_{:k}^{(t)}-W_{:k})^\top Z\Pi|| ||Z^\top W^{(t)}||\\
         &\lesssim\frac{K^{1.5}\sqrt{p_{max}}\lt}{n^{1.5}\Delta_{min}}.
    \end{align*}
     Furthermore, by the same argument as before, 
     \begin{align*}
          || (W_{:k}^{(t)}-W_{:k})^\top EW^{(t)}||&\leq || (W_{:k}^{(t)}-W_{:k})^\top E(W^{(t)}-W)||+ || (W_{:k}^{(t)}-W_{:k})^\top EW||\\
          &\lesssim K||E||\max_k||W_{:k}-W_{:k}^{(t)}||^2+\frac{K^2\sqrt{p_{max}}}{n^{1.5}\Delta_{min}^2} l(z^{(t)},z)\\
          &\lesssim \frac{K^2\sqrt{p_{max}}}{n^{1.5}\Delta_{min}^2} l(z^{(t)},z).
     \end{align*}
    We obtain the result by triangle inequality.
    
    \paragraph{Proof of 5.} Since $Z^\top W =I_K$ we have \begin{align*}
        ||Z^\top W^{(t)}|| &\leq 1 + || Z^\top (W^{(t)}-W)||\\
        & \lesssim 1 + ||I - Z^\top W^{(t)}||\\
        & \lesssim 1+K\frac{\lt}{n\Delta^2_{min}} \tag{by Equation \eqref{eq:izw}}\\
        &\lesssim 1 \tag{by assumption on $\tau^{(0)}$}.
    \end{align*}
\end{proof}

\begin{lemma}\label{lem:lambdat}
For \sircsbm\, we have with probability at least $1-n^{-\Omega(1)}$ that for all  $z^{(t)}$ such that $\lt \leq \tau^{(0)}$
\[ \max_{k\in [K]}|n_k^{(t)}-n_k| \leq \frac{ l(z^{(t)},z)}{\Delta_{min}^2}, \quad |\lambda^{(t)}-\lambda|\lesssim \lambda \frac{K\lt}{n\Delta_{min}^2}.\]
\end{lemma}
\begin{proof}

First observe that \[ \max_{k\in [K]}|n_k^{(t)}-n_k| = \max_{k\in [K]}|\sum_i \indic_{\lbrace Z_i^{(t)}=k\rbrace}-\indic_{\lbrace Z_i=k\rbrace}|\leq h(z^{(t)},z) \leq \frac{\lt}{\Delta_{min}^2}  \] by Lemma \ref{lem:ht}. 

Then note that we have \begin{align*}
    |p_{max}^{(t)}-p_{max}| &\leq  \max_{k,k'} ||(W^{(t)}_{:k})^\top AW^{(t)}_{:k'} -W_{:k}^\top PW_{:k'}|| \tag{the max is 1-Lipschitz}\\
                    & \leq \max_{k,k'}(||(W^{(t)}_{:k})^\top EW^{(t)}_{:k'} ||  + ||(W^{(t)}_{:k}-W_{:k})^\top PW^{(t)}_{:k'} ||+||W_{:k}^\top P(W^{(t)}_{:k'}-W_{:k'} )|| )\\
                    & \lesssim \max_{k}||W^{(t)}_{:k} ||^2||E|| + \frac{K^{1.5}\sqrt{p_{max}}\lt}{n^{1.5}\Delta_{min}} \tag{by the proof of Lemma \ref{lem:wt}}\\
                    &\lesssim \frac{K\sqrt{p_{max}}}{\sqrt{n}} + \frac{K^{1.5}\sqrt{p_{max}}\lt}{n^{1.5}\Delta_{min}}\\
                    &\lesssim \frac{K^{1.5}\sqrt{p_{max}}\lt}{n^{1.5}\Delta_{min}} \\
                    &\lesssim \sqrt{\frac{K}{np_{max}}}\frac{K\lt}{n\Delta_{min}}p_{max}\\
                    &\lesssim \frac{K\lt}{n\Delta_{min}^2}p_{max}
\end{align*} 
since $\Delta_{min} \asymp \sqrt{\log n/K} \asymp \sqrt{np_{max}/K}  $. Consequently,
\begin{align*}
    \abs{\frac{\lambda^{(t)}}{\lambda}-1} &\leq \abs{\frac{n_{min}^{(t)}}{n_{min}}\frac{p_{max}}{p_{max}^{(t)}}-1} \\
    &\leq \abs{\frac{n_{min}^{(t)}-n_{min}}{n_{min}}\frac{p_{max}}{p_{max}^{(t)}} } + \abs{\frac{p_{max}-p_{max}^{(t)}}{p^{(t)}_{max}}}\\
    &\lesssim \frac{K \lt}{n\Delta_{min}^2}  + \frac{K\lt}{n\Delta_{min}^2}\\
    &\lesssim \frac{K\lt}{n\Delta_{min}^2}.
\end{align*}
\end{proof}
\begin{lemma}
\label{lem:lambdat_ssbm}
For \ircssbm, we have  with probability at least $1-n^{-\Omega(1)}$ that for all  $z^{(t)}$ such that $\lt \leq \tau^{(0)}$ \[ |\lambda^{(t)}-\lambda|\lesssim \lambda \frac{K\lt}{n\Delta_{min}^2}\]
\end{lemma}
\begin{proof}
By a similar argument as the one used in Lemma \ref{lem:lambdat} we have with probability at least $1-n^{-\Omega(1)}$ that
\[ |p^{(t)}-p| \lesssim \frac{K\lt}{n\Delta_{min}^2}p, \quad |q^{(t)}-q| \lesssim \frac{K\lt}{n\Delta_{min}^2}q.\]
This implies  
\begin{align*}
    \frac{p^{(t)}-q^{(t)}}{p-q} = 1 + O\left(\frac{(p+q)K\lt}{(p-q) n\Delta_{min}^2}\right)  
     = 1 +  O\left(\frac{K\lt}{n\Delta_{min}^2}\right) 
\end{align*} 
because $p-q\gtrsim p$. Thus, 
\begin{align*}
    \left|\log\left(\frac{p^{(t)}}{q^{(t)}}\right)-\log\left(\frac{p}{q}\right)\right| = \left|\log\left(\frac{p^{(t)}}{p}\frac{q}{q^{(t)}} \right)\right|
    = 2\log\left(1+O\left(\frac{K\lt}{n\Delta^2_{min}}\right) \right)
     = O\left(\frac{K\lt}{n\Delta^2_{min}}\right).
\end{align*}
Hence 
\[ \left|\frac{ \log(\frac{p^{(t)}}{q^{(t)}})}{\log(\frac{p}{q})} -1\right| = O\left(\frac{K\lt}{n\Delta^2_{min}}\right)\] 
since $\log(p/q)$ is bounded above by assumption.
Consequently, 
\[ \frac{\lambda^{(t)}}{\lambda}-1=O\left(\frac{K\lt}{n\Delta_{min}^2}\right).\]
\end{proof}

\section{Additional experiments}\label{app:xp}
Section \ref{app:xp_sbm} presents experimental results of our method applied on non-assortative SBM when the covariates are disregarded. Section \ref{app:xp_rainfall} contains additional results for the Australia rainfall data for different choices of $K$. Section \ref{app:xp_sponge} includes additional results for signed SBM and Section \ref{app:not_well} contains an experiment when each source of information by itself does not allow to separate the communities. Finally, in Section \ref{app:rd_init}, we study empirically the robustness of our algorithm to random initialization.

\subsection{Heterophilic SBM}\label{app:xp_sbm}
If we disregard the covariates, our algorithm can be used for inference under a general SBM, in contrast to the method proposed by \cite{Lu2016StatisticalAC} which was restricted to the assortative setting. In particular, our algorithm also works for networks with heterophilic communities. The following experiment illustrates the gain in term of accuracy for \ircsbm \, initialized with \asc \ (spectral clustering on the adjacency matrix). It also shows the interest of using more than one iteration in the refinement step with the MAP (this corresponds to \irmapp). 

\begin{figure}[H]
    \centering
    \includegraphics[scale=0.6]{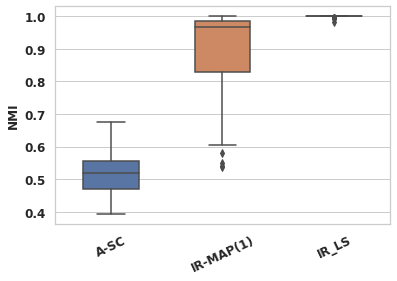}
    \caption{ Average  performance  of  different  algorithms on a heterophilic SBM, over $40$ Monte Carlo runs.}
    \label{fig:heter_sbm}
\end{figure}

We consider $n=1000, K=3$, $Z_i \overset{i.i.d}{\sim }\text{Multinomial}(1; 1/3,1/3,1/3)$ and 
\[ \Pi = \begin{pmatrix}
                        0.2 & 0.05 & 0.1 \\
                        0.05 & 0.15 & 0.05 \\
                        0.1 & 0.05 & 0.03 
                      \end{pmatrix}.
\]
The NMI is averaged over $40$ repetitions; the results are shown in Figure \ref{fig:heter_sbm}. We also considered the VEM algorithm implemented in the R package \texttt{blockmodels} \citep{r2016Blockmodels}, but the running time was prohibitive (approximately one hour for a single Monte Carlo run, whereas our algorithm take a few seconds). It nevertheless returned the exact partition as \ircsbm.

\subsection{Australian Rainfall}\label{app:xp_rainfall}
\begin{figure}[H]
    \centering
    \includegraphics[scale=0.5]{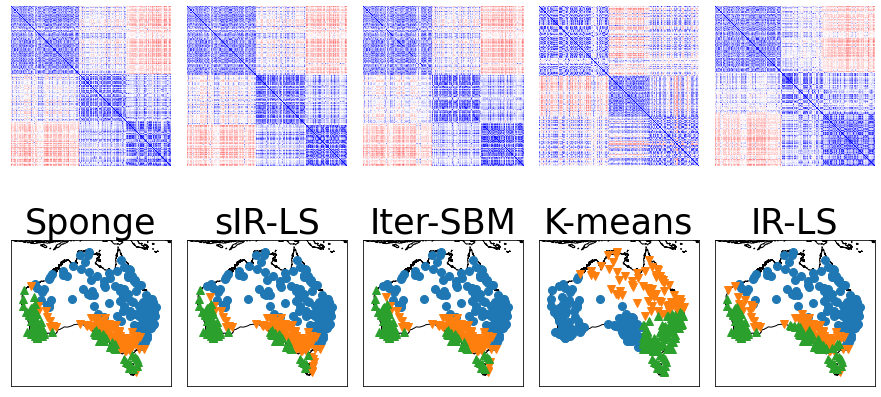}
    \caption{ Sorted adjacency matrices of the Australian rainfall data set and corresponding maps for $K=3$.}
    \label{fig:ausk3}
\end{figure}

\begin{figure}[H]
    \centering
    \includegraphics[scale=0.5]{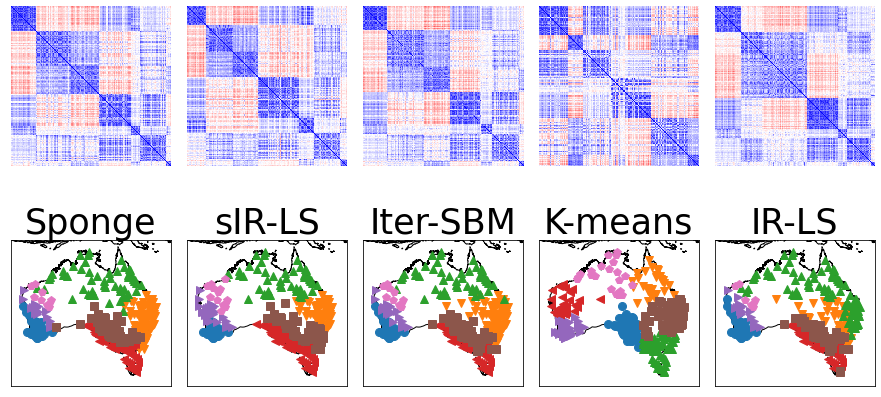}
    \caption{ Sorted adjacency matrices of the Australian rainfall data set and corresponding maps for $K=7$.}
    \label{fig:ausk7}
\end{figure}

\begin{figure}[H]
    \centering
    \includegraphics[scale=0.5]{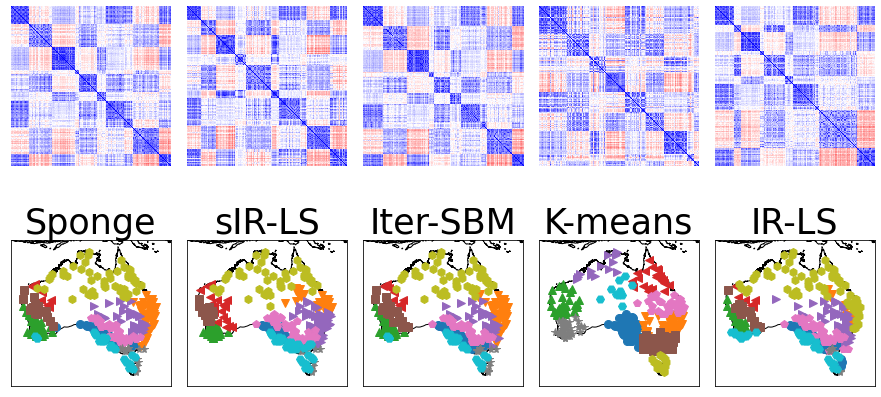}
    \caption{ Sorted adjacency matrices of the Australian rainfall data set and corresponding maps for $K=10$.}
    \label{fig:ausk10}
\end{figure}

We reproduced the experiment presented in Section \ref{sec:xp} but with $K \in \set{3,7,10}$ (see Figures \ref{fig:ausk3}, \ref{fig:ausk7} and \ref{fig:ausk10}). We can observe that \sircsbm\, provides a visibly better clustering than that by \ircsbm. This can possibly be attributed to the fact that \ircsbm\, requires estimating more parameters than \sircsbm, however this dataset is quite small in size. Furthermore, the partition provided by \kmeans\, is visibly different than that generated by \sircsbm, \spongesym \, and \irsbm \, (although there is still some overlap) since \kmeans\, uses only covariate information. %
%

\subsection{Signed SBM}\label{app:xp_sponge}
\begin{figure}[H]
    \centering
    \includegraphics[scale=0.7]{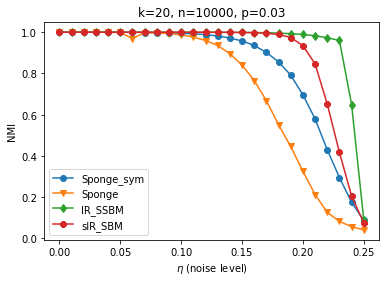}
    \caption{ NMI for varying noise $\eta$, and with $p = 0.03$, $K= 20$, $n=10000$.}
    \label{fig:signed}
\end{figure}

We reproduced the experiment presented in Section \ref{sec:xpsigned} at a different sparsity level $p=0.03$. The relative performance of the methods (shown in Figure \ref{fig:signed}) are similar to that in Figure \ref{fig:sponge}. Algorithm \ref{alg:irssbm} outlines our iterative refinement method (namely \irssbm) for clustering signed graphs, under the Signed SBM. It seems that there is a threshold for the noise level $\eta$ above which no algorithm can succeed. We conjecture that \irssbm\, is optimal and attains this threshold.

\begin{algorithm}[H]
\caption{\irssbm}\label{alg:irssbm}
\begin{flushleft}
        \textbf{Input:} The number of communities $K$, initial partition $z^{(0)}$, $T\geq 1$.
\end{flushleft}
        \begin{algorithmic}[1]
        \FOR{$0\leq t\leq T-1$} 
            \STATE Compute $W^{(t)}=Z^{(t)}(D^{(t)})^{-1}$ where $D^{(t)}= \diag (n_k^{(t)})_{k\in [K]}$, and $C^{(t)}=AW^{(t)}$.
            \STATE Update the partition for each $i\leq n$  \[  z_i^{(t+1)}=\arg\max_k C_{ik}^{(t)}\] 
        \ENDFOR
        \end{algorithmic}
 \textbf{Output:} A partition of the nodes $z^{(T)}$.
\end{algorithm}

\subsection{Not distinguishable community}\label{app:not_well}
\begin{figure}[H]
    \centering
    \includegraphics[scale=0.7]{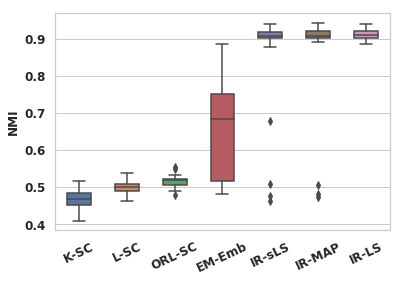}
    \caption{ Performance of different algorithms on CSBM. Results are sorted by mean NMI and obtained over $40$ runs.}
    \label{fig:boxplot2}
\end{figure}

We repeated the experiment of Section \ref{subsec:csbm_not_well_sep} with the rank deficient connectivity matrix 
\[ \Pi = 0.02*\begin{pmatrix}
                        1.5 & 1.5 & 0.05 \\
                        1.5 & 1.5 & 0.05 \\
                        0.05 & 0.05 & 1.5 
              \end{pmatrix}
\] 
but with the same covariate parameters as in Section \ref{subsec:csbm_not_well_sep}. Not surprisingly, we obtained similar results (see Figure \ref{fig:boxplot2}) as in Figure \ref{fig:boxplot}. The main difference is that the performance of \orlsc\, worsened. 

\subsection{Random initialization and performances under the threshold for exact recovery.}\label{app:rd_init}

In this experiment we fix $K=2$ and $n=1000$. Let $c>0$ and define $p=4c\log n/n$ and $q=c\log n/n$. Also define a Gaussian mixture in $\R$ with centers $1$ (for community $1$) and $2$ (for the second community). The variance parameter $\sigma$ is chosen such that the information provided by the Gaussian part of the model is equal to the one provided by the graph. More precisely, we set $\sigma^2=\frac{1}{2c\log n}$. The value $c=0.5$ corresponds to the threshold for exact recovery.

\begin{figure}[H]
    \centering
    \includegraphics[scale=0.7]{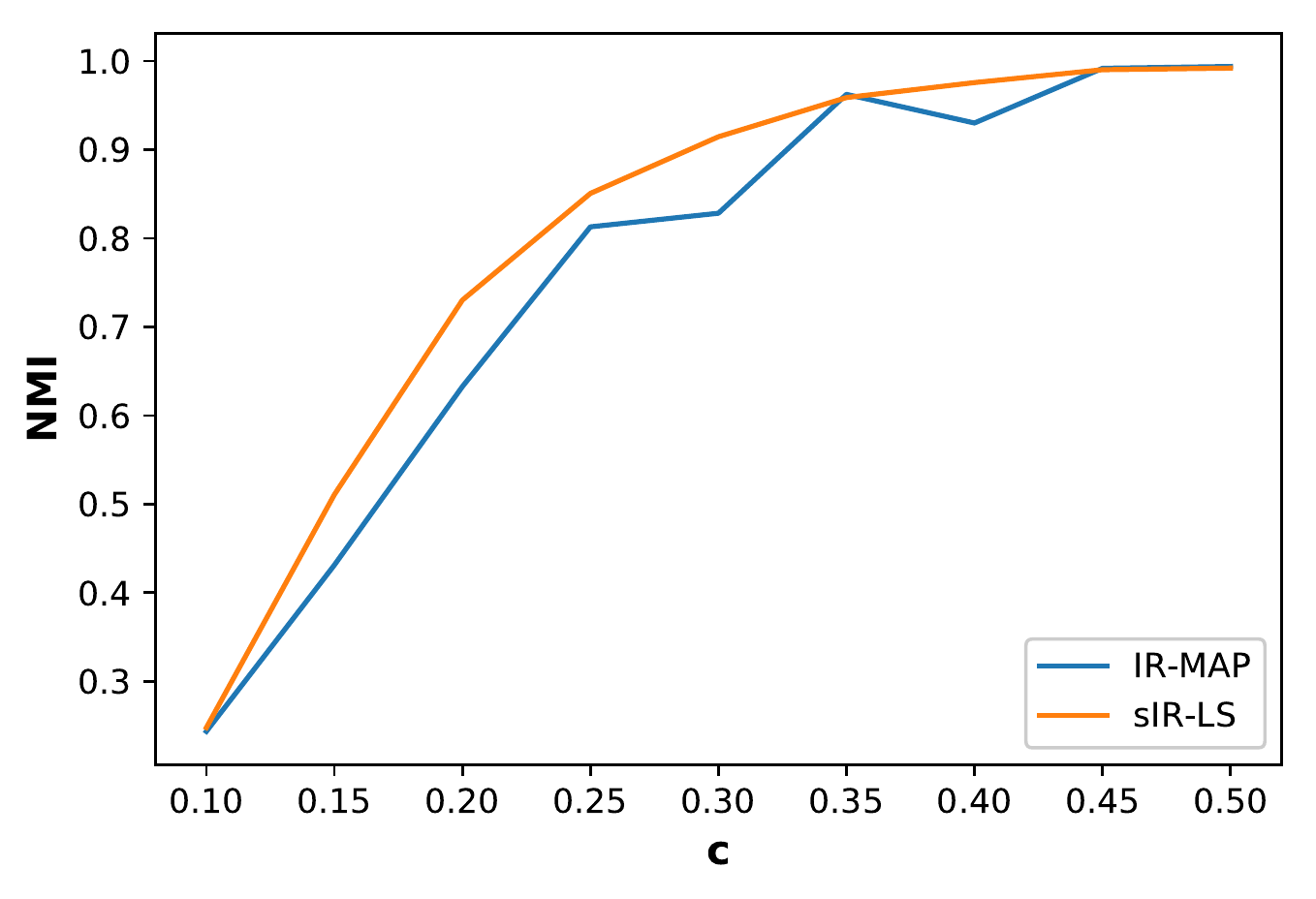}
    \caption{ Average performance measured by NMI obtained with random initialization over $20$ runs.}
    \label{fig:xp_rd_init}
\end{figure}

Figure \ref{fig:xp_rd_init} shows that when initialized with a random $Z^{(0)} \overset{i.i.d}{\sim }\text{Multinomial}(1; 1/2,1/2)$, \sircsbm\, performs slightly better than \irmap\,, has less variability, and is robust to random initialization, hence justifying the Gaussian approximation.

\subsection{Comparison between random initialization and initialization with \gmm}
We use a similar experimental setting as the one described in Section 5.1. We only slightly change the connectivity matrix  \[ \Pi = 0.02*\begin{pmatrix}
                        1.6 & 1.2 & 0.5 \\
                        1.2 & 1.6 & 0.5 \\
                        0.5 & 0.5 & 1.2 
                      \end{pmatrix}.
\]

\begin{figure}[!ht]
    \centering
    \includegraphics[scale=0.55]{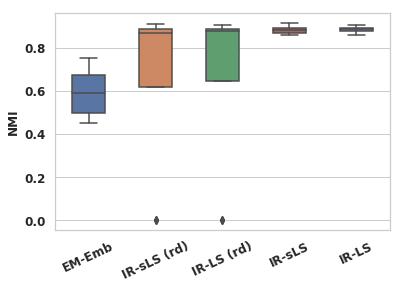}
    \caption{Average performance over $20$ runs of our algorithms under the experimental setting of Section $5.1$. Here (rd) indicates random initialization.}
    \label{fig:rd_vs_gmm}
\end{figure}

Figure \ref{fig:rd_vs_gmm} shows that when randomly initialized ours algorithms \ircsbm\, and \sircsbm\, can suffer from numerical instability. That's why we recommend to use \gmm\, instead. But it could interesting to develop a strategy based on random initialization by identifying and disregarding the random initialization that lead to atypical results.

\end{document}